\documentclass[acmtog, authorversion]{acmart}

\AtBeginDocument{%
  \providecommand\BibTeX{{%
    \normalfont B\kern-0.5em{\scshape i\kern-0.25em b}\kern-0.8em\TeX}}}

\setcopyright{acmcopyright}
\acmJournal{TOG}
\acmYear{2021} \acmVolume{41} \acmNumber{1} \acmArticle{1} \acmMonth{7} \acmPrice{15.00}\acmDOI{10.1145/3470848}

\usepackage{graphicx}
\usepackage{subcaption}
\usepackage{float}
\usepackage{xcolor}
\usepackage{multirow}
\newcommand{\revised}[1]{\textcolor{black}{#1}}
\newcommand{\minorrev}[1]{\textcolor{black}{#1}}

\begin{document}

\title{\revised{SofGAN:} A Portrait Image Generator with Dynamic Styling}

%
\author{Anpei Chen}
\authornote{Authors contributed equally to this work.}
\email{chenap@shanghaitech.edu.cn}
\orcid{1234-5678-9012}
\author{Ruiyang Liu}
\authornotemark[1]
\email{liury@shanghaitech.edu.cn}
\affiliation{%
  \institution{ShanghaiTech University}
  \city{Shanghai}
}

\author{Ling Xie}
\affiliation{%
 \institution{ShanghaiTech University}
 }
\email{xieling@shanghaitech.edu.cn}

\author{Zhang Chen}
\affiliation{%
 \institution{ShanghaiTech University}
 }
\email{chenzhang@shanghaitech.edu.cn}

\author{Hao Su}
\affiliation{%
  \institution{University of California San Diego}
  }
\email{myemail@shanghaitech.edu.cn}

\author{Jingyi Yu}
\affiliation{%
  \institution{ShanghaiTech University}
  \city{Shanghai}
}
\email{yujingyi@shanghaitech.edu.cn}

\authorsaddresses{Author's addresses:
Anpei Chen, Ruiyang Liu, Zhang Chen, Ling Xie, Jingyi Yu, Shanghai Engineering Research Center of Intelligent Vision and Imaging, School of Information Science and Technology, ShanghaiTech University, Shanghai, China; Anpei Chen, Ruiyang Liu, and Zhang Chen are also affiliated with the Shanghai Institute of Microsystem and Information Technology (SIMIT) and the University of the Chinese Academy of Science (UCAS); Hao Su, Department of Computer Science and Engineering, University of California San Diego, La Jolla, CA, USA.
}



\begin{abstract}

Recently, \minorrev{\textit{Generative Adversarial Networks} (GANs)} have been widely used for portrait image generation. However, in the latent space learned by GANs, different attributes, such as pose, shape, and texture style, are generally entangled, making the explicit control of specific attributes difficult. To address this issue, we propose a \revised{$SofGAN$} image generator to decouple the latent space of portraits into two subspaces: a geometry space and a texture space. The latent codes sampled from the two subspaces are fed to two network branches separately, one to generate the 3D geometry of portraits with canonical pose, and the other to generate textures. The aligned 3D geometries also come with semantic part segmentation, encoded as a \textit{semantic occupancy field (SOF)}. The SOF allows the rendering of consistent 2D semantic segmentation maps at arbitrary views, which are then fused with the generated texture maps and stylized to a portrait photo using our \textit{semantic instance-wise (SIW)} module.
Through extensive experiments, we show that our system can generate high quality portrait images with independently controllable geometry and texture attributes. The method also generalizes well in various applications such as appearance-consistent facial animation and dynamic styling. The code is available at \href{https://apchenstu.github.io/sofgan/}{\textcolor{blue}{sofgan.github.io.}}
\end{abstract}



\begin{teaserfigure}
  \includegraphics[width=\textwidth]{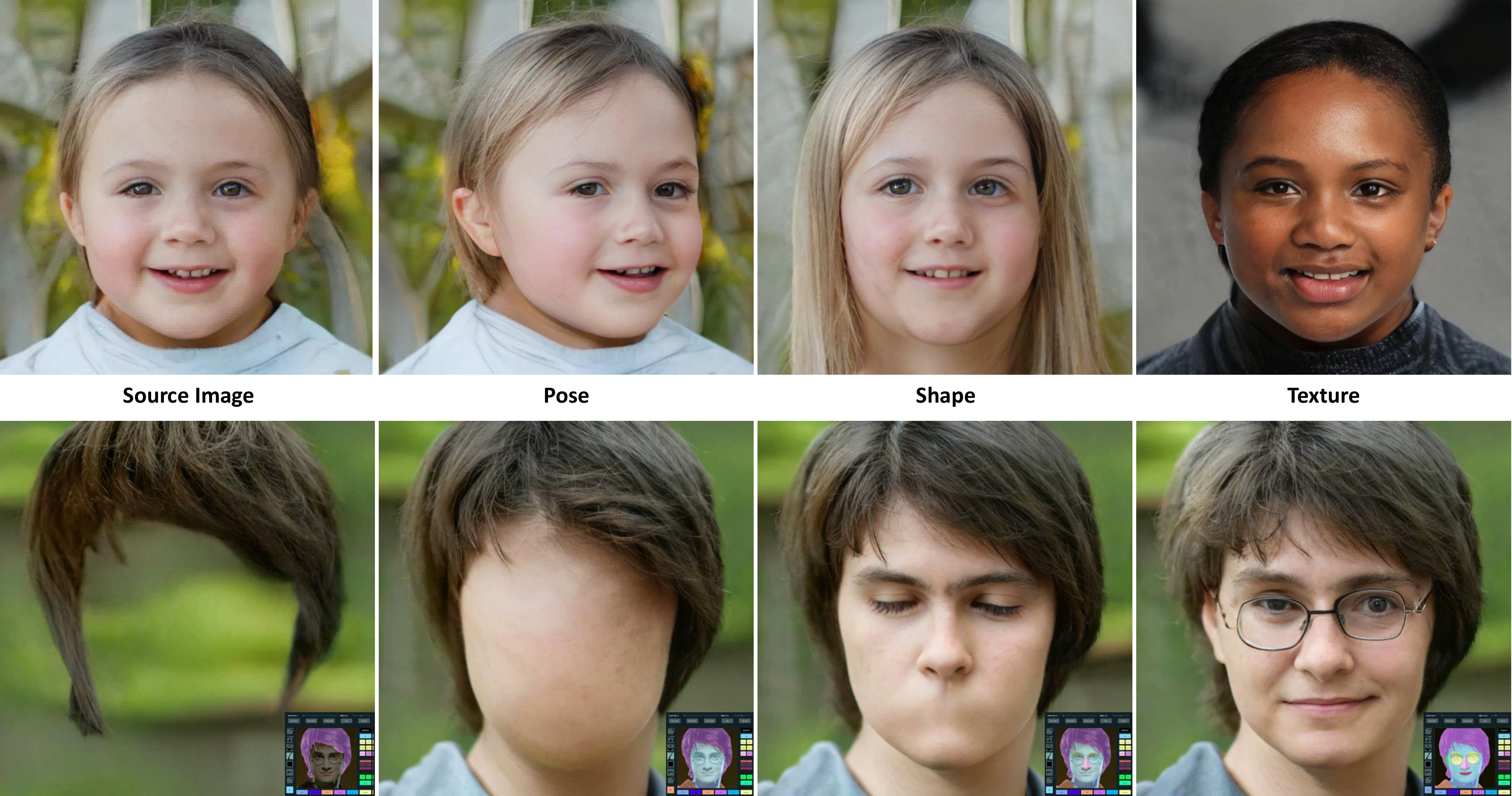}
  \caption{First row: \minorrev{our portrait image generator allows explicit control over pose, shape and texture styles}. Starting from the source image, we explicitly change it's head pose (2nd image), facial/hair contour (3rd image) and texture styles. Second row: interactive image generation from incomplete segmaps. Our method allow users to gradually add parts to the segmap and generate colorful images on-the-fly.}
  \label{fig:teaser}
\end{teaserfigure}

\maketitle

\section{Introduction}

\begin{figure*}[ht]
\begin{center}
   \includegraphics[width=1.0\linewidth]{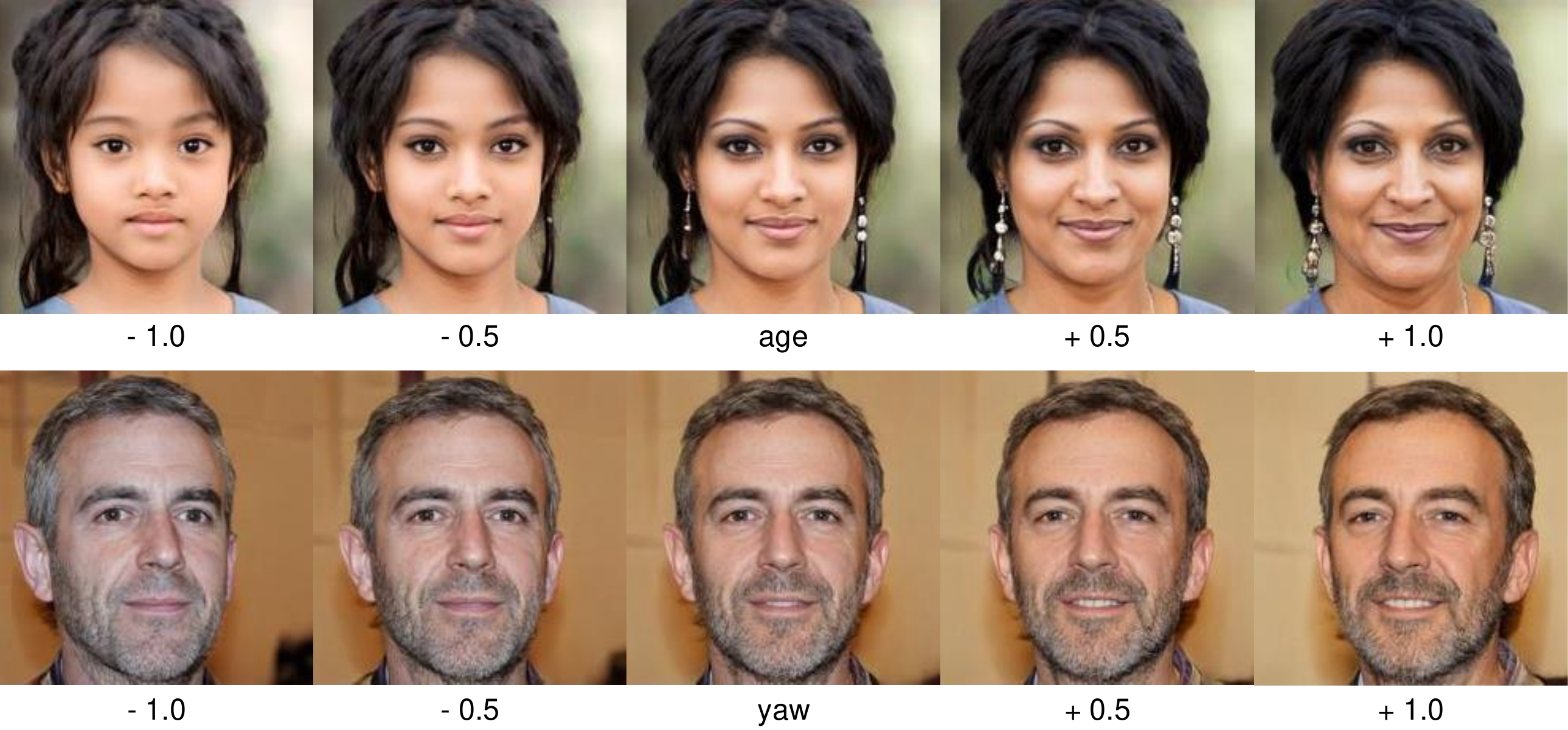}
\end{center}
\caption{Visualization of entangled attributes in the generation space of StyleGANs. In the first row, when we change the age \minorrev{attribute} of the face, other attributes like hair style and ear rings are also changed. In the second row, complexion and facial emotions are changed along with yaw pose.}
\label{fig:direction}
\end{figure*}

Portrait modeling and rendering find broad applications in computer graphics, ranging from virtual reality (VR) and augmented reality (AR), to photorealistic face swaps, and to avatar-based tele-immersion. Effective portrait generators should be capable of synthesizing a diverse set of styles at high realism with explicit controls over attributes including illumination, shape (e.g., expression, poses, wearings), texture styles (e.g., age, gender, hair, and clothes), etc.
    
\revised{Physically-based rendering approaches have sought to explicitly model shape, materials, lighting, and textures and can potentially provide users complete controls over these attributes for synthesizing various styles. In reality, most existing techniques require carefully processed models as well as extensive manual tuning to achieve desired photorealism. More recent approaches have adopted learning-based schemes that aim to model styles as specific image distributions from a large-scale image dataset. Their major advantage is the avoidance of explicitly formulating geometry, lighting, or material that generally requires extensive experience and knowledge. For example, methods based on Generative Adversarial Networks (GAN) \cite{karras2019analyzing,karras2019style} can now achieve extremely high realism and variety; however, there is still ample room to improve controllability over styles. }

\revised{The seminal attribute-based approaches \cite{chen2016infogan,shen2020interfacegan,tewari2020stylerig,abdal2020styleflow} explore to find attribute-specific paths in the latent spaces, e.g., to use direction codes to drag random style code towards each attribute direction. Such methods assume that attributes are linearly interpolative and independent from each other in the pre-trained generation space. However, this assumption is generally violated because attributes are entangled in the latent space, leading to flickering and undesirable attribute change.}

\revised{It is also possible to employ 3D priors ~\cite{tewari2020stylerig,Deng2020DisentangledAC} in portrait synthesis. These techniques can produce  convincing facial animations and relighting results under user control. However, limited by the representation ability of explicit 3D facial priors (most commonly, the 3D morphable model), such approaches fail to model decorative attributes such as hairstyles, clothing, etc. The recent \textit{Semantic Region-Adaptive Normalization (SEAN)} technique ~\cite{zhu2020sean} provides regional style adjustment conditioned on a segmentation map for more flexible image synthesis and editing. However, it also requires boundary-aligned segmentation maps and images (i.e., paired) for training that are difficult to acquire in practice. }

We propose a disentangled portrait generation method by drawing inspirations from the practices of highly successful image rendering systems. In the image rendering literature, it is a basic practice to decompose the modeling of scenes as constructing the geometry component and the texture component. Likewise, we learn two individual latent spaces, i.e., a geometry space and a texturing space. Sparked by recent works on implicit geometry modeling \cite{mescheder2019occupancy,Sitzmann2019SceneRN,park2019deepsdf,chen2019learning}, we extend the traditional occupancy field to \textit{semantic occupancy field (SOF)} to model portrait geometry. \textit{SOF} describes the probabilistic distribution over $k$ semantic classes (including hair, face, neck, cloth, etc.) for each spatial point. We train this \textit{SOF} from calibrated multi-view semantic segmentation maps without groundtruth 3D supervision. To synthesize images, we first raytrace the \textit{SOF} to obtain 2D segmentation maps from a given user-specified viewpoint then adopt GAN generator to texture each semantic region with a style code sampled from the texturing space. We propose a \textit{semantic instance-wise (SIW)} texturing module to support dynamic and regional style control. Specifically, we design a novel semantic-wise ``demodulation'' and a novel training scheme that spatially mixes two random style codes for each semantic region during training. We further encode the semantic segmentation maps in a low dimensional space with a three-layer encoder to encourage continuity during view changing.

We have evaluated our method on the FFHQ dataset~\cite{karras2019style} and the CelebA dataset~\cite{CelebAMask-HQ}. Our generator achieves a lower Fréchet Inception Distance (FID score) and higher Learned Perceptual Image Patch Similarity (LPIPS) metric than the SOTA image synthesis methods. We will release our code, pre-trained models, and results upon acceptance.   

To summarize, we propose a photorealistic portrait generator (\textit{SofGAN}) with the following characteristics:

\begin{itemize}
	\item \textbf{Attribute-consistent style adjustment.} Our new representation provides explicit controls over individual attributes with the rest unchanged and hence can support respective rendering effects such as free-viewpoint rendering, global and regional style adjustment, face morphing, expression editing, and artificial aging.
	
	\item \textbf{Training on unpaired data.} Unlike previous approaches that require using paired/aligned RGB and segmentation images for training, our SIW module can be directly trained using unpaired real-world images and synthesized semantic segmentation maps.

	\item \textbf{On-demand and Interactive Generation.} The tailored architecture of our generator supports photorealistic portrait synthesis from inaccurate or even incomplete segmentation maps. We have hence built a user-friendly tool that allows users to hand draw semantic contours for interactive portrait design (see the supplementary video).
\end{itemize}

\section{Related Work}

\textbf{Unconditional image generation.} In the context of photo-realistic image generation, the pioneering ProgressiveGan \cite{karras2017progressive} conducts progressive training on both generator and discriminator and achieves significant improvements on rendering quality, training efficiency, and stability. Subsequent StyleGAN and extensions \cite{karras2019style,karras2019analyzing} re-design the generator architecture to provide scale-specific control: the generator starts from a learned constant block and adjusts the “style” at each convolution layer. Such a structure enables direct controls over image features at different scales, i.e., unsupervised separation of high-level attributes from stochastic variations (e.g., freckles, hair) to support intuitive scale-specific mixing and interpolation. By providing a less entangled representation, these approaches produce state-of-the-art (SOTA) results.

\textbf{Conditional image generation}, also commonly referred to as style transfer, leverages GANs to learn a mapping from the source domain to the target domain (\cite{isola2017pix2pix,wang2018pix2pixHD,park2019SPADE,zhu2020sean}). Pix2Pix \cite{isola2017pix2pix} firstly adopts a U-Net decoder in the generator to directly share the low-level information between input and output through skip-connections. It then trains the network with both perceptual loss and GAN loss. To tackle instability in adversarial training as well as to handle high-resolution image generation, the subsequent Pix2PixHD \cite{wang2018pix2pixHD} uses a multi-scale generator and discriminator architectures. Although effective, these approaches face vanishing/exploding gradients when using a deep network. SPADE \cite{park2019SPADE} proposes to use spatially adaptive normalization on all decoding layers rather than the first few layers of the network. The recent SEAN \cite{zhu2020sean} aims to synthesize style specified images by combining style latent vector and semantic maps, and achieves SOTA \textit{Frechet Inception Distance} (FID) score. Since both SPADE and SEAN use the perceptual loss \cite{simonyan2014very} based on paired training data, their generated images still present certain blurriness.

\begin{figure*}[ht]
\begin{center}
   \includegraphics[width=1.0\linewidth]{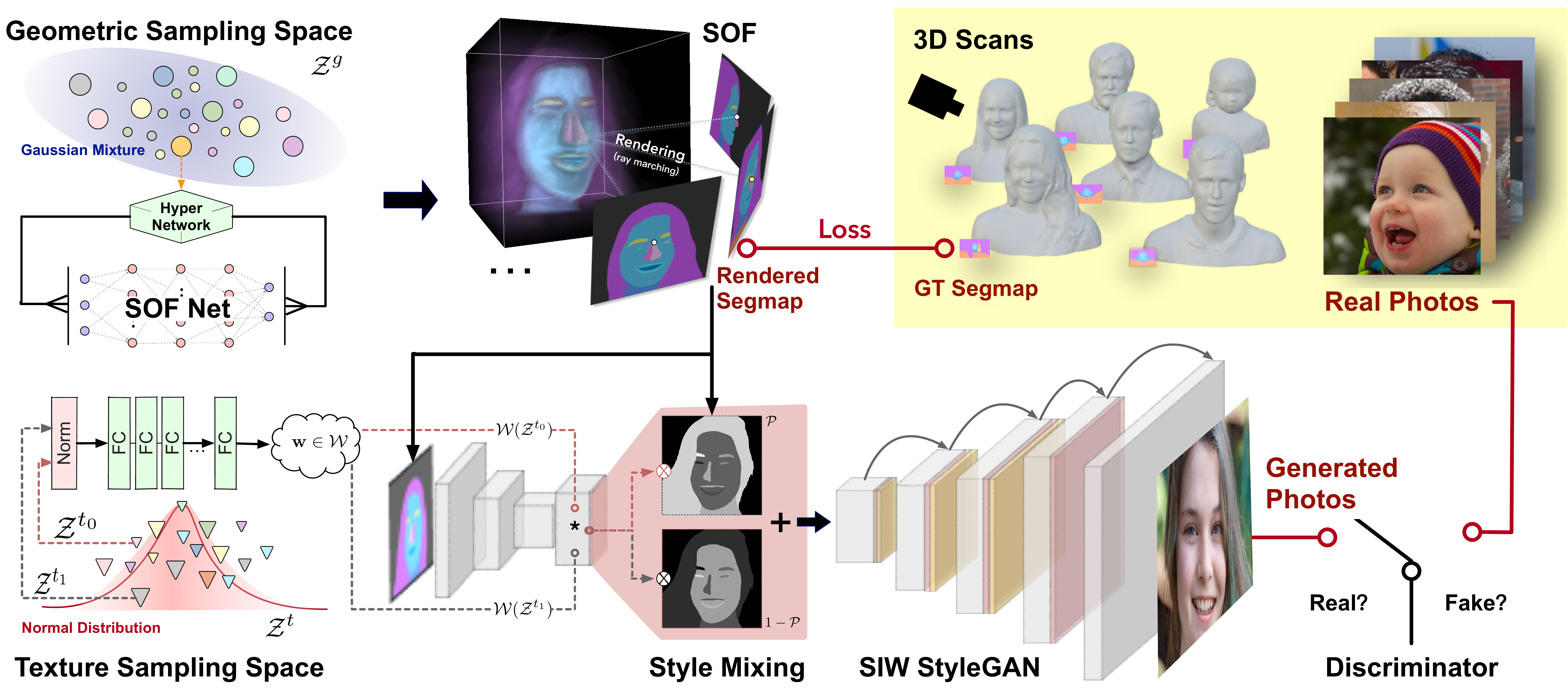}
\end{center}
\caption{An overview of our generation pipeline. \minorrev{Geometric latent} \revised{code $z^g$ in a learned Gaussian mixture. We adopt a hyper network to decode $z^g$ into the weights of a MLP (\textit{SOF Net}) which represents a contiguous semantic occupancy \minorrev{field} (\textit{SOF}) among 3D space. Then we render free-viewpoint segmentation maps from \textit{SOF} with a ray-casting-marching scheme} and generate two region-wise distance maps $\mathcal{P}, 1.0-\mathcal{P}$ for the following texturing.\revised{ During texturing, we sample random vectors ($z^t_0, z^t_1$) from the texture space and mod/demod into per-layer style vectors}, then we adopt a Semantic Instance Wise (SIW) StyleGAN module to regionally stylize the generated segmaps. We use $5$ style mixing layers (Pink) and $3$ SPADE layer (Brown) in the SIW-StyleGAN. Our SOF and SIW are trained separately, where we render multi-view segmentation maps from synthetic portrait scans to train SOF and use real photos for SIW-StyleGAN training.}
\label{fig:pipeline}
\end{figure*}

\textbf{Controllable image synthesis.} Unconditional GANs unanimously synthesize images from randomly sampled latent vectors and thus are incapable of providing attribute-specific control (e.g., pose, eye, age). Inspired by the disentangled representation in the VAE and GAN literature, several recent approaches explore to achieve controllable image synthesis by disentangling the generation space into semantically meaningful attributes, either based on vector arithmetic phenomenon \cite{shen2019interpreting, Collins20ganlocal} or shape priors \cite{tewari2020stylerig} such as 3DMM \cite{blanz1999morphable}. Other controllable neural modeling methods \cite{park2019deepsdf,chen2019learning,mescheder2019occupancy, nguyen2019hologan, nguyen2020blockgan} employ traditional implicit geometry representations such as the signed distance field (SDF) and occupancy field (OF) but under neural network approximations. In a similar vein, controllable neural rendering techniques extend physical-based models to conduct dynamic image generation tasks such as free-viewpoint relighting \cite{chen2020neural}, reenactment \cite{geng2018warp,thies2016face2face,siarohin2019first,tewari2020stylerig,cao2013facewarehouse}, novel views synthesis \cite{thies2019deferred,zhou2020rotate,chen2018deep, mildenhall2020nerf, lombardi2019neural, sitzmann2019deepvoxels, Sitzmann2019SceneRN}, etc, even with sparse inputs.

\revised{Instead of using explicit SDF and occupancy-based neural scene representations, our approach models the shared coarse geometry across different portraits via an embedded semantic occupancy implicit fields. Specifically, we set out to model the SOF using a set of semi-calibrated segmentation maps and subsequently encoding for the distribution of each 3D point over $k$ semantic classes. This allows us to obtain free-viewpoint segmentation maps by projecting SOF onto a 2D space under a specific view camera. We further use a novel SIW module for modeling texture/appearance on each region in the semantic map. The combination of SOF and SIW modules enables better disentanglement between geometry attributes and texture styles, thus providing a new flexible control over image generation while maintaining generation quality.}

\section{Overview}

\revised{Our portrait generator synthesizes photo-realistic portrait images with controls over attributes including shape, pose, and texture styles. In StyleGAN, styles are controlled via features at each Conv layer based on the input latent vectors at different levels: coarse (contour), medium (expressions, hairstyle) and fine levels (color distribution, freckles). Although effective, such a control mechanism does not provide independent \revised{controls over individual attributes, largely }due to the entanglement of various attributes. }

To address this issue, we decompose the generation space into two sub-spaces: a geometry space and a texture space (Fig. ~\ref{fig:pipeline} left). Each sample in the geometry space can be decoded into the weights of a SOF Net that represents a 3D-continuous occupancy field (\textit{SOF}) with companion semantic labelings.  At the rendering stage, given an arbitrary query viewpoint, we use a ray marching framework to map the \textit{SOF} to a 2D segmentation map. The use of SOF ensures view consistency. Next we follow the semantic image synthesis framework \cite{isola2017pix2pix,zhu2017unpaired,wang2018pix2pixHD,park2019SPADE,zhu2020sean} and present a semantic-based, instance-wise (\textit{SIW}) generation module to generate photorealistic images.

\revised{In a nutshell, our complete image generation process can be formulated as:}

\begin{equation}
\begin{aligned}
I = G(\mathcal{W}(z^t), \mathcal{R}(SOF(z^g), C)),
\label{eq:SOF_sep}
\end{aligned}
\end{equation}
\revised{where $z^g\in\mathbb{R}^n$ denotes the geometric code, $z^t\in\mathbb{R}^m$ the texture code, and $m$ and $n$ the dimensions of the geometric and texture latent spaces respectively. $SOF$ corresponds to a neural geometric representation defined by $z^g$ whereas $\mathcal{R}$ a differentiable renderer to render $SOF$ onto a segmentation map at the query viewpoint $C$. $G$ refers to the \textit{SIW}-StyleGAN image generator and $\mathcal{W}$ is a basis transformation operator on the texture code.}

\revised{It is worth mentioning that a major limitation of previous condition-GAN style transfer methods is the requirement of using pairwise content/style images for training. In practice, collecting such image pairs is not only time-consuming but also requires elaborate \minorrev{alignment}. Our technique does not impose such constraints. Rather, we set out to separately train \textit{SOF} and \textit{SIW}. In particular, training a SOF only requires using a relatively small set of multi-view segmentation maps that can even originate from scanned 3D models (Section ~\ref{geometry modeling with SOF}). In sec.\ref{sec:siw stylegan}, we present SIW-StyleGAN to support regional texturing from segmentation maps.}

\revised{\section{Geometry Modeling}
\label{geometry modeling with SOF}
Previous approaches use the signed distance function (\textit{SDF}) or the occupancy field (\textit{OF}) as an alternative geometric representation to 3D points or meshes. For example, the SDF function $\mathcal{F}(x)$  defined on $\mathbb{R}^3$ models the signed distance value or occupancy probability. Such functions only characterize surface geometry (spatial location, normal, etc..) but do not consider the semantic meanings of the surface. We observe that such semantic components are critical for photo-realistic image generation: they provide crucial guidance on generating convincing texture and shading details. }

\revised{We instead use a novel semantic occupancy field (\textit{SOF}) representation that extends the SDF by mapping an input point $x \in \mathbb{R}^3$ to a $k$-dimensional vector to describe the occupancy probability of it belonging to  $k$ different semantic classes (e.g., eyes, mouth, hair, hat, clothing, etc.). The combination of semantic labels with geometry in SOF have several key benefits. }

\begin{enumerate}
    \item Regions defined by semantic classes serve as the basic units for regional style synthesis and adjustment. Such regional controls over individual components of the portrait (e.g. hair vs. eyes vs. mouth) greatly outperform the brute-force approaches that treat an image as a single entity.
    \item \revised{Free-viewpoint image generation is a natural extension of SOF that can be seamlessly integrated into the deep networks: SOF describes the geometry attributes where new views can be directly synthesized via subsequent neural rendering. }

\end{enumerate}

\revised{An \textit{SOF} models both geometry and semantic labels as:
\begin{equation}
\begin{aligned}
    &\mathcal{F}: \mathbb{R}^3 \to \mathbb{R}^k, \quad \mathcal{F}(x)=P_x
\end{aligned}
\label{eq:sof_fmt}
\end{equation}
where the semantic occupancy function \minorrev{$\mathcal{F}(x)$} assigns  to every point $x \in \mathbb{R}^3$ in 3D space a $k$-dimensional probability ($P_x$) over $k$ semantic classes.}

\begin{figure}[t]
\begin{center}
   \includegraphics[width=0.95\linewidth]{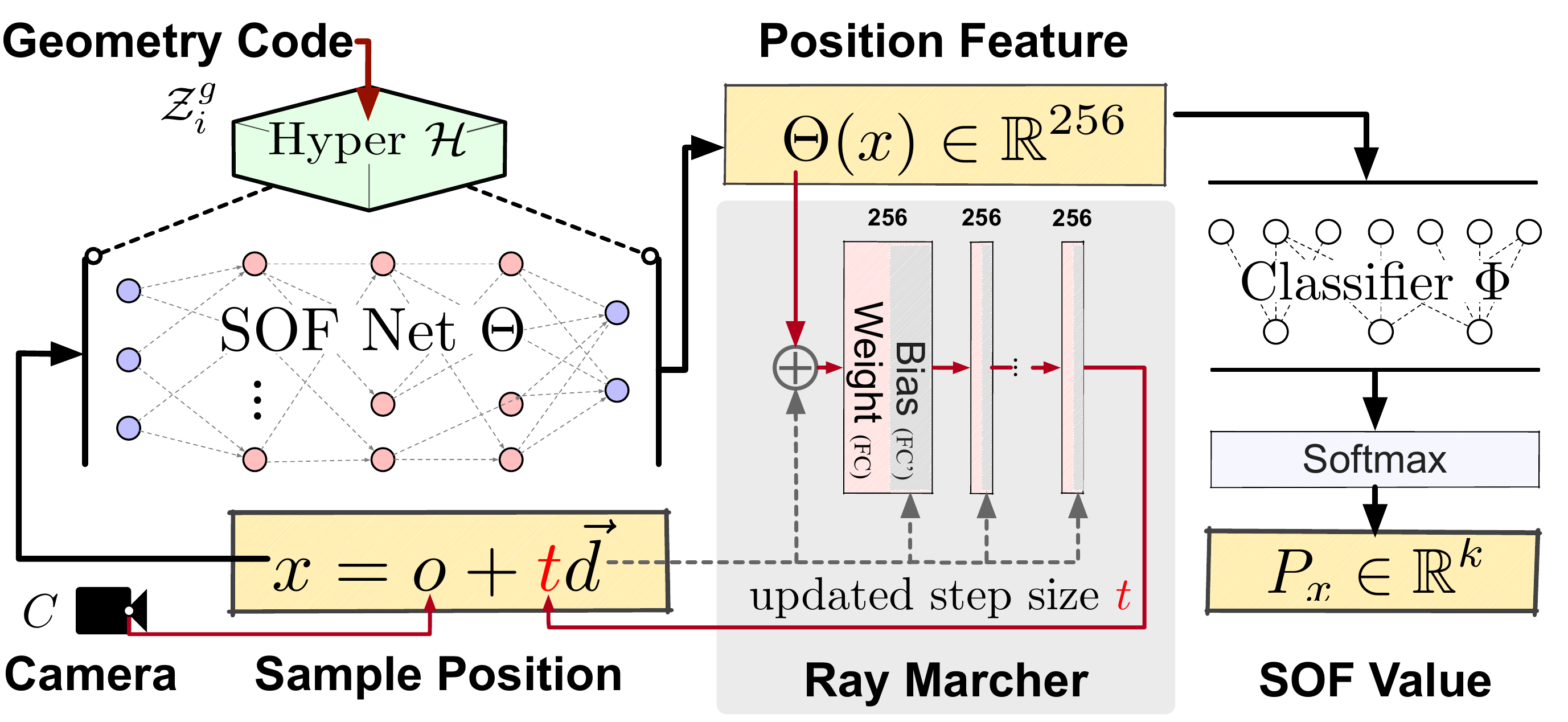}
\end{center}
\caption{\revised{The free viewpoint segmaps rendering process. Given a query camera $C$ and geometric latent code $z^{g}_{i}$, we first use a hypernet $\mathcal{H}$ to compute the weight of SOF network ($\Theta$). Next, we conduct ray casting on SOF. To render each pixel, we cast its corresponding ray towards the SOF, starting from the given camera's Center-of-Projection (CoP) \minorrev{$o$}. We conduct iterative depth refinement using the SOF where the refinement step is also computed from an MLP-based ray marcher. Finally, we feed the estimated surface point and its feature vector to the classifier $\Phi$ to decode them into a $k-$class semantic probability $P_s\in\mathbb{R}^k$.}}
\label{fig:sof_architecture}
\end{figure}

\subsection{Neural SOF Representation} 
\label{SOF Network Approximate}
\revised{Inspired by recent advances on neural scene representations ~\cite{Sitzmann2019SceneRN,Oechsle2020LearningIS,Saito2019PIFuPI,Peng2020ConvolutionalON,park2019deepsdf}, we approximate $\mathcal{F}$ using two multi-layer perceptrons (MLPs), an SOF net $\Theta$, and a classifier $\Phi$. As shown in Fig. \ref{fig:sof_architecture}, $\Theta$ maps each spatial location $x\in\mathbb{R}^3$ to a spatial feature vector $f\in\mathbb{R}^n$ that encodes the semantic property whereas $\Phi$ corresponds to a semantic classifier with softmax activation to decode $f$ to a $k$-dimensional semantic probability vector $P_x$ as:}
 
\begin{equation}
\begin{aligned}
    P_x \approx \Phi(\Theta(x)) \quad where \quad \Theta: \mathbb{R}^3 \to \mathbb{R}^n, \ \Phi: \mathbb{R}^n \to \mathbb{R}^k
\label{eq:SOF_sep}
\end{aligned}
\end{equation}


\revised{We train the \textit{SOF} by rendering known 3D portrait models into multi-view segmentation maps (segmaps) and compute the binary cross entropy loss between rendered segmaps and ground truth segmaps. The ground truth \minorrev{segmentation} maps can be obtained by either semantically parsing real multi-view portrait images or by rendering synthetic 3D models with semantic labels (e.g., on textures), as shown in Fig. \ref{fig:sof_train_data} of the Appendix. }

\begin{figure}[t]
\begin{center}
   \includegraphics[width=\linewidth]{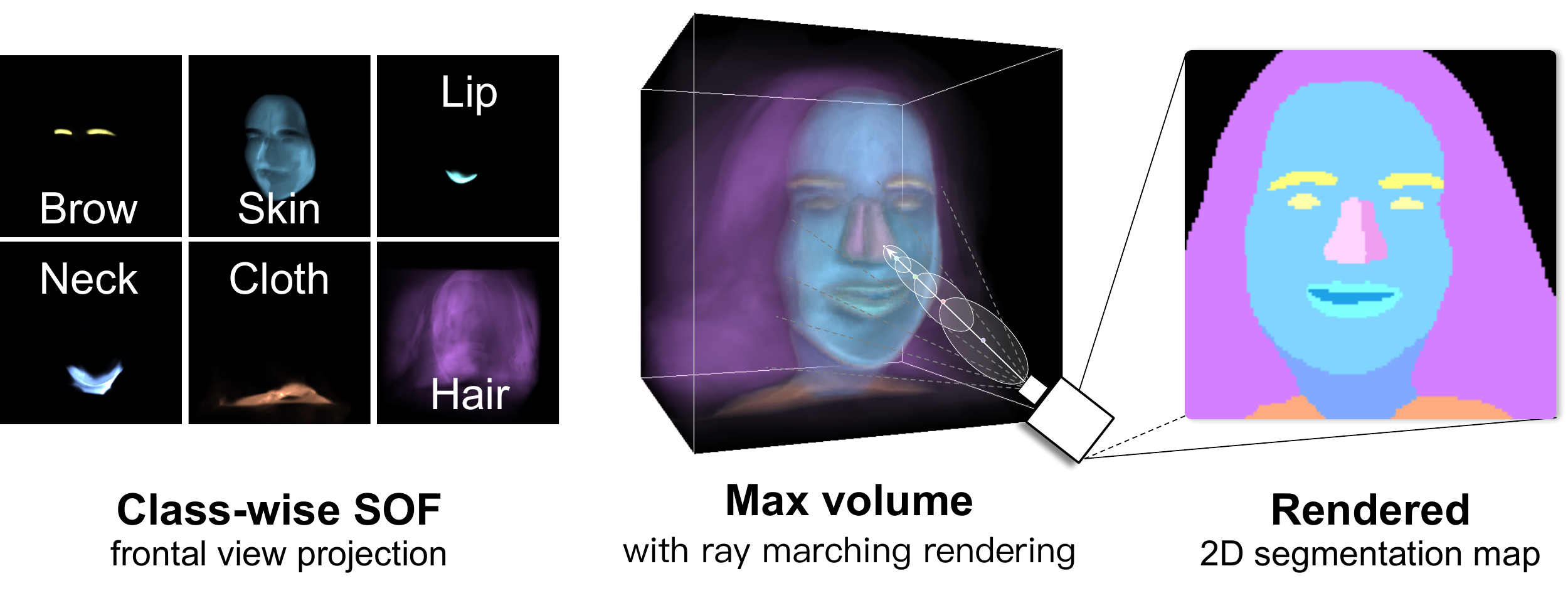}
\end{center}
\caption{\revised{\textit{SOF} is a $3-$dimensional volume, with a $k-$class semantic probability for each spatial point. The probability for each semantic class also forms a contiguous 3D volume (left). For each position, we take the semantic value with max probability and adopt a differentiable ray marching scheme (middle) to render it into 2D segmentation maps (right).}}
\label{fig:SOF}
\end{figure}

\revised{To tackle diversity across 3D portraits, the brute-force approach would be to train a separate SOF for every portrait instance. Such approaches are highly inefficient and thus impractical for real-world applications. We instead train a geometric sampling space to support various portrait instances in the latent space using latent codes. We observe that 3D human portraits exhibit similar layouts, i.e., facial components such as eyes, hair, nose, and lips follow consistent spatial layouts. This enables us to train a shared canonical latent geometry space. Specifically, we adopt a shared hyper-network $\mathcal{H}$ (Fig. \ref{fig:sof_architecture}) to learn, for each instance's network, the weights of \textit{SOF} net$\Theta$. Under the shared hyper-network, we can represent each instance in the training dataset with a latent vector $z^g$ as:}

\begin{equation}
    \mathcal{H}: \mathbb{R}^m \to \mathbb{R}^{\|\Theta\|}, \quad \mathcal{H}(z^{g}_i)=\Theta_i,
    \quad i \in \{1,...,\|\mathcal{D}\|\},
\label{eq:hyper}
\end{equation}
where $i$ refers to the $i$th instance in training dataset $\mathcal{D}$. We use all $z^{g}$s to form our geometry sampling space $\mathcal{G}$ with the Gaussian mixture model. In our implementation, we build a dataset $\mathcal{D}$ with $122$ segmented portrait scans and for each model we render 64 segmaps from randomly sampled viewpoints. 
\revised{
\subsection{Free-Viewpoint Segmap Rendering}
 Given the trained SOF, we can subsequently render free-viewpoint segmaps. Fig. \ref{fig:sof_architecture} shows our detailed rendering procedure: we extract the portrait surface from the SOF via a ray marching scheme and estimate the per-ray depth $t$ (from the camera center to the surface) as the sum of $N$ marching step-sizes $t_i, i \in [1,N]$. The portrait surface $\mathcal{S}$ corresponds to points:}
\revised{
\begin{equation}
\mathcal{S} = \{x \in \mathbb{R}^3 \ | \ x=o + \vec{d}  \sum_{i=1}^N t_i\}
\end{equation}
}
\revised{
For each forward rendering pass, we first \minorrev{cast} rays to the scene, and then use a ray marching scheme to obtain the surface point $x$ (middle of Fig. \ref{fig:SOF}) by iteratively estimating the marching step-size $t_i$, and finally predict its semantic property with the Eq.\ref{eq:SOF_sep}. The multiply steps in the ray marching can be simulated with a $LSTM$ network\cite{Sitzmann2019SceneRN}, and achieves pleasing scene representation performance. However, we observe that the $LSTM$ structure is quite unstable and sensitive to novel views during viewpoint changing, as shown in Fig. \ref{fig:ablation_sof} of the Appendix.}

\revised{
We thus propose a more stable ray marcher (Fig. \ref{fig:sof_architecture}, middle) that predicts the step sizes from the current position feature and the ray direction. Each ray marching MLP layer contains two sub-fully-connected layers $FC_i$ and $FC^{'}_i$ where $FC_i$ is a linear mapping without additive bias (i.e. the $b$ in linear layer $y=xA^T+b$) whereas $FC^{'}_i$ maps the ray direction to be the layer bias. }

\begin{equation}
\begin{aligned}
    &f_{i} = ReLU(FC_i(f_{i-1})+FC^{'}_i (\vec d)), \\
    &i\in[1,...,6],\quad f_0=\Theta(x)
\label{eq:ray_marching}
\end{aligned}
\end{equation}

\revised{
Fig. \ref{fig:SOF} shows a sample SOF with the predicted semantic labels (we only show the label that corresponds to the highest probability) for each point within the volume. We also demonstrate the improvement in the rendering quality of our proposed MLP ray marching module over the LSTM one in Fig. \ref{fig:ablation_sof} of the Appendix. To further ensure depth consistency across views at the inference stage, we extract a coarse portrait proxy with marching cube algorithm (MC) and use the reprojected depth as the initial depth of ray marching.}


\section{Texture Synthesis}
\label{sec:siw stylegan}

\revised{Next, we show how to use the 2D semantic segmaps from an arbitrary viewpoint to synthesize photorealistic images. Our approach is to conduct image-to-image translation \cite{isola2017pix2pix,zhu2017unpaired,wang2018pix2pixHD,park2019SPADE,zhu2020sean} and we present a \textit{SIW-StyleGAN} technique that support high image quality and flexible style adjustments at both global and local scales. }

\revised{
To achieve attribute-specific generation, existing StyleGANs~\cite{karras2019style,karras2019analyzing} start synthesizing from a learnable $4\times4\times512$ constant block and upsample it to higher resolutions ($4^2-1024^2$) conditioned on the style vector $z^t$. Specifically, we transform the style vector to its corresponding kernel scales and conv image features $F_{in}$ at each convolution layer where the output of each layer $F_{o}$ can be formulated as:}

\begin{figure}[t]
  \includegraphics[width=\linewidth]{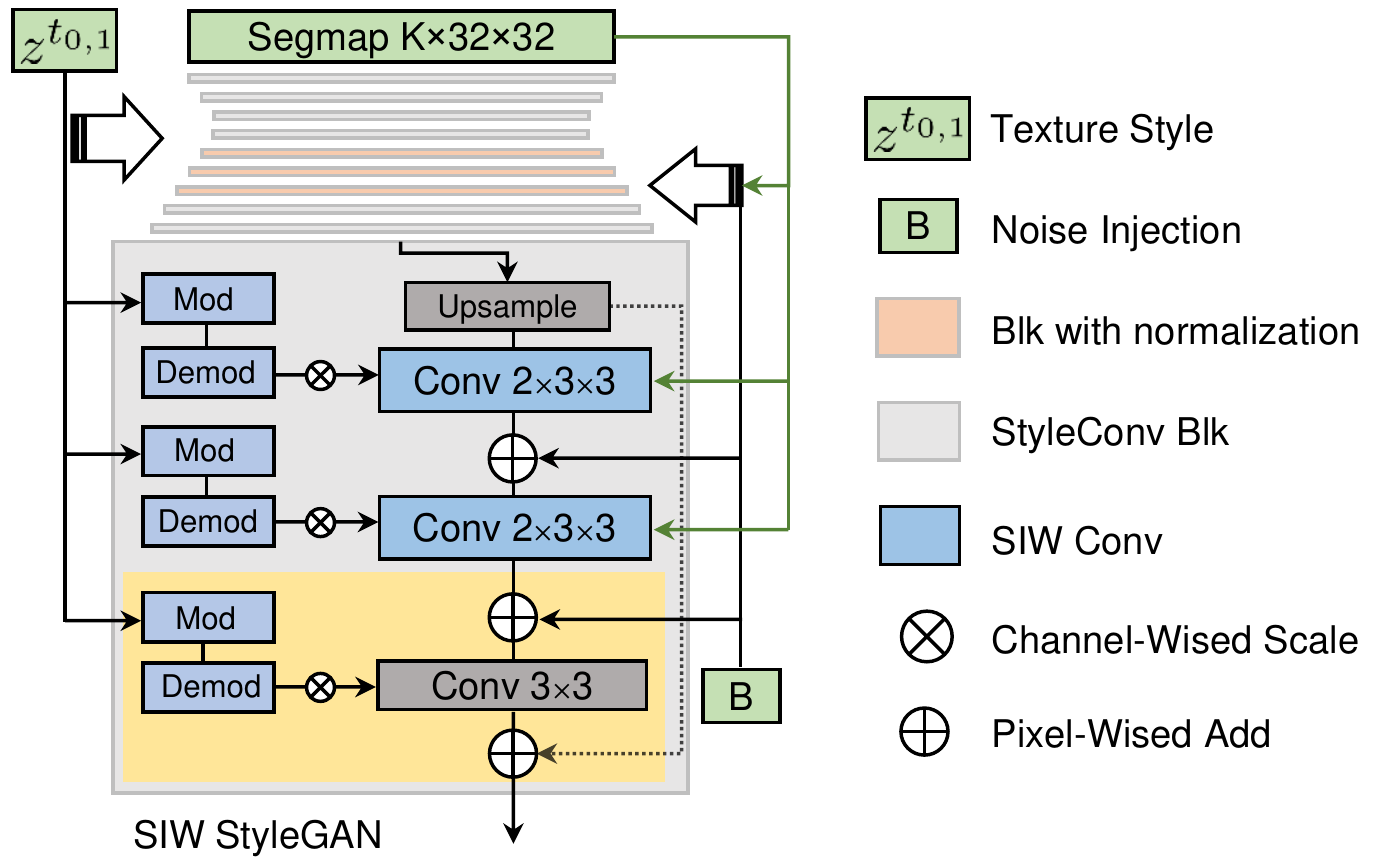}
  \caption{Our SIW generation module. SIW extends StyleGANs2 to generate images from one-hot segmap and two normal distributed $z^{t_{0,1}}$, \revised{ where $z^{t_i}$ stands for texture style code}. "Mod/Demod" refers to the $\Psi$ term in Eq. \ref{eq:stylegan_conv} that transforms the style vectors to the style code $\mathcal{W}(z^t_{0,1})$ \minorrev{after Mod operation}. We formulate the SIW Conv operation as Eq.\ref{eq:stylegan_conv} and adopt SPADE normalization layer \cite{park2019SPADE} at $64^2-256^2$ resolutions (i.e., the orange blocks). Details of each layer can be found in Appendix A.}
  \label{fig:net_generator}
\end{figure}

\revised{
\begin{equation}
\label{eq:stylegan_conv}
F_{o} = F_{in}  \ast w^{'}\quad \text{with} \quad w^{'} = \alpha \cdot w \text{ and } \alpha = \Psi(\mathcal{W}(z^t)),
\end{equation}
where $w$ and $w^{'}$ are the original and the modulated convolution kernels, respectively, $\ast$ is the convolution operator, and $\alpha \in \mathbb{R}^{1 \times CH}$ is a channel-wise feature scaling vector obtained by feeding the style vector $\mathcal{W}(z^t)$ to the modulation/demodulation function $\Psi$. $z \sim \mathcal{N}(0,1)$ is the input of StyleGANs sampled from a $512$ dimensional normal distribution with mean $0$ and variance $1$.   }

\revised{
To fully exploit the semantic segmaps from our scheme, we further formulate the texturing process as to regionally stylize the one-hot semantic segmap $\mathcal{M} \in \mathbb{R}^{K \times 32 \times 32}$ with a vector $z^t \in \mathbb{R}^{512}$ of normal distribution. We first use three convolutional layers to downscale $\mathcal{M}$ to a low-resolution feature map. Next we use a semantic instance-wise (SIW) convolution layer to conduct region-based upscale of the feature map to a high-resolution image (as the SIW StyleConv block in Fig. ~\ref{fig:net_generator}). We achieve the SIW style convolution operation by applying a region-specific convolution. Specifically, we first extend the modulation into $K$ (i.e., the number of semantic labels, we set to $17$ in all our examples) style vectors and then convert it into a pixel- and channel-wise scaling map, thus the output feature maps of each layer are formulated as the sum of all regional feature maps:}
\revised{
\begin{equation}
\begin{aligned}
F_{o} &= \sum^{K}_{i=1} (F_{in} \ast w^{'}_{i}) \cdot  \mathcal{M}_i \\
w^{'}_{i} =& \alpha_{i} \cdot w_{i}; \quad \alpha_{i} = \Psi(\mathcal{W}(z^{t}_i))
\label{eq:conv}
\end{aligned}
\end{equation}
where $\alpha \in \mathbb{R}^{K \times CH}$ includes the scaling factors for each feature channel and semantic region. }

\revised{
To strengthen the influence of the semantic segmaps in the generation process, we add an SPADE layer ~\cite{park2019SPADE,zhu2020sean} in the intermediate scaling layers (i.e., $64^2-256^2$ \minorrev{resolution} shown as the orange blocks in Figure~\ref{fig:net_generator}) and we have:}

\revised{
\begin{equation}
F_{o} = \gamma \sum^{K}_{i=1} (F_{in} \ast w^{'}_{i}) \cdot  \mathcal{M}_i + \beta\\
\label{eq:conv_spade}
\end{equation}
where  $\beta$ and $\gamma$ are the mean and variance of the spatially adaptive normalization parameters: $\gamma, \beta = SPADE(\mathcal{M})$. }

\revised{
Recall that the operation above requires convolving the feature block with $K$ different style kernels before fusing the intermediate feature blocks into one feature block (e.g., the summing in Eq.\ref{eq:conv_spade}). Brute-force implementation is neither efficient nor necessary: the one-hot semantic segmap contains mostly zeros so that most intermediate features do not contribute to the fused feature block.}

\revised{Note that, the modulated regional texture styles $w^{'}$ are generated from random vectors in the training phase, we hence approximate the generation by spatially mixing two style vectors $z^{t_{0,1}}$ with a region-wise distance map $\mathcal{P} \in [0,1]$ (i.e., the mixed style training scheme)},  where the distance represents the similarity between the two styles. Each semantic region maps to a single style distance so that the output features can be simplified to a \minorrev{regional linear blending}, the linear combination of two modulated features. Eq. \ref{eq:conv_spade} is then simplified to:

\begin{equation}
F_{o} =  \gamma \cdot (F_{in} \ast \mathcal{W}(z^{t_0}) \cdot \mathcal{P} 
+ F_{in} \ast \mathcal{W}(z^{t_1}) \cdot (1.0 - \mathcal{P})) + \beta
\label{eq:conv_new}
\end{equation}

At the training stage, we obtain the distance map $\mathcal{P}$ by assigning a random value to each semantic region. This mixed-style scheme simplifies $K$ Conv and addition operations to $2$ and therefore supports highly efficient training.

\begin{table*}[ht]
\begin{center}
\begin{tabular}{c|c|c|c|c|c|c|c}
\hline\hline
 & \multicolumn{2}{c|}{\textbf{INPUTS}} & \multicolumn{2}{c|}{\textbf{TRAINING}}  & \multicolumn{3}{c}{\textbf{EFFECTS}}\\ \cline{2-8}
                  & latent code & segmentation & pairwise & 3D segmaps & global style  & local style  & free view    \\ \hline
\textbf{Pix2PixHD}\cite{wang2018pix2pixHD}         & $\surd$ & $\surd$ & $\surd$ & & & &\\ \hline
\textbf{SPADE}\cite{park2019SPADE}             & $\surd$ & $\surd$ & $\surd$ & & $\surd$ &  &\\ \hline
\textbf{SEAN}\cite{zhu2020sean}              & $\surd$ & $\surd$ & $\surd$ & & $\surd$ & $\surd$ &\\ \hline
\textbf{StyleGAN2}\cite{karras2019analyzing}        & $\surd$ & & & & $\surd$ & &\\ \hline
\textbf{SofGAN}      & $\surd$ & $\surd$ &  & $\surd$ & $\surd$ & $\surd$  & $\surd$ \\ \hline
\end{tabular}
\end{center}
\caption{\revised{Comparisons on training data requirement and controls over styles of our vs. SOTA methods. Pix2pix ~\cite{wang2018pix2pixHD}, SPADE ~\cite{park2019SPADE}, and SEAN ~\cite{zhu2020sean} unanimously require a large number of pair-wise (image and its registered segmap) training data whereas ours requires multi-view segmaps obtained from a small number of 3D scans. On controls, our technique provides more explicit and more varieties of controls than StyleGANs.}}
\label{tb:comp_input}
\end{table*}

\begin{figure*}[ht]
\begin{center}
   \includegraphics[width=0.95\linewidth,height=0.28\linewidth]{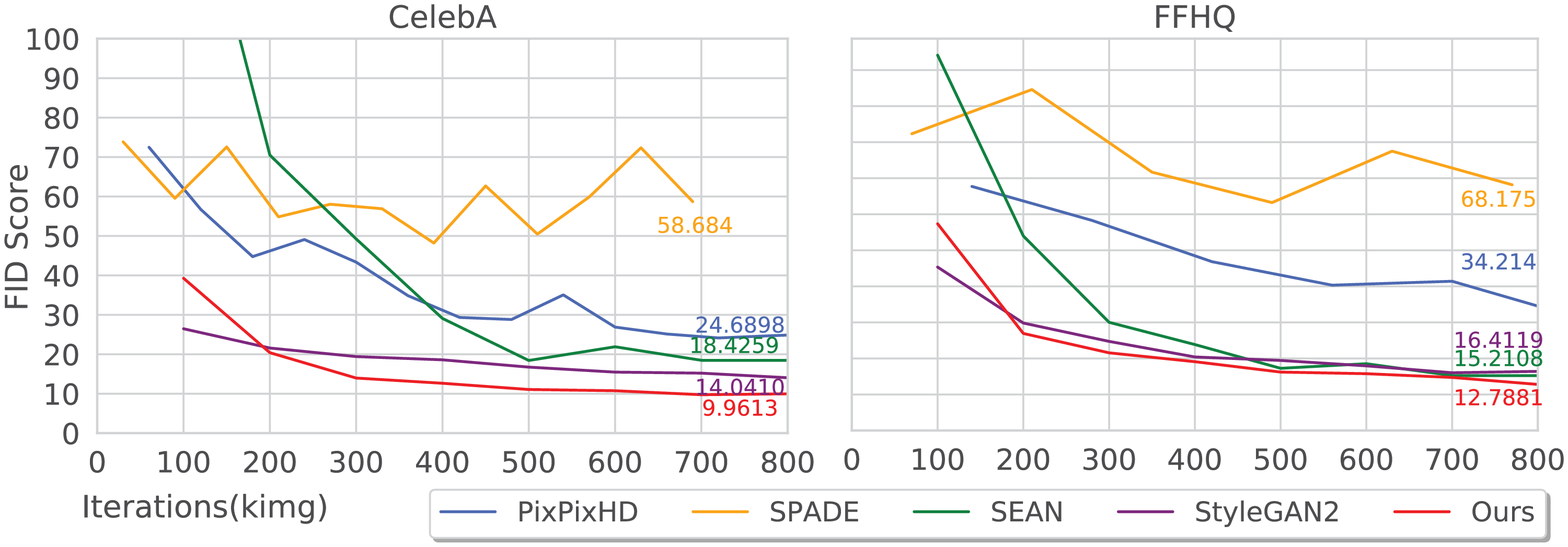}
\end{center}
   \caption{Quantitative comparisons between our method vs. SOTA: we compare FID scores on CelebA\cite{CelebAMask-HQ} and FFHQ\cite{karras2019style}. Specifically, we first re-train all networks on one dataset and then generate images with random style vectors on other datasets for evaluation.}
\label{fig:curve}
\end{figure*}

\definecolor{Blue}{rgb}{0.00,0.00,1.00}
\begin{table*}[ht]
\begin{center}
\begin{tabular}{c|ccc|ccc}
\toprule\hline
 & \multicolumn{3}{c|}{\textbf{CelebA}}  & \multicolumn{3}{c}{\textbf{FFHQ}} \\ \cline{2-7}
& AlexNet $\uparrow$ & VGG 16 $\uparrow$  & SqueezeNet $\uparrow$ & AlexNet $\uparrow$  & VGG 16 $\uparrow$   & SqueezeNet $\uparrow$\\ \hline
\textbf{Pix2PixHD}\cite{wang2018pix2pixHD}  & $0.63_{\pm 7.38}$ & $0.60_{\pm 4.92}$   & $0.45_{\pm 6.93}$ & $0.53_{\pm 8.27}$ & $0.55_{\pm 6.04}$ & $0.35_{\pm 6.71}$ \\ 

\textbf{SPADE}\cite{park2019SPADE}  & $0.52_{\pm 8.27}$   & $0.53_{\pm 6.48}$   & $0.36_{\pm 6.93}$  & $0.49_{\pm 10.51}$ & $0.51_{\pm 8.50}$ & $0.34_{\pm 9.17}$  \\ 

\textbf{SEAN}\cite{zhu2020sean}   & $0.51_{\pm 12.52}$ & $0.57_{\pm 11.40}$ & $0.41_{\pm 12.30}$    & $0.58_{\pm 6.26}$   & $0.66_{\pm 6.04}$  & $0.50_{\pm 6.71}$ \\ 

\textbf{StyleGAN2}\cite{karras2019analyzing}  & $0.62_{\pm 7.38}$   & $0.62_{\pm 7.38}$   & $0.47_{\pm 7.16}$   & $0.64_{\pm 6.04}$ & $0.66_{\pm 5.59}$ & $0.51_{\pm 6.48}$ \\ 

\textbf{SofGAN}   & $\textbf{0.65}_{\pm 6.71}$ & $\textbf{0.66}_{\pm 6.04}$ & $\textbf{0.50}_{\pm 7.16}$ & $\textbf{0.66}_{\pm 5.81}$ & $\textbf{0.69}_{\pm 5.59}$ & $\textbf{0.53}_{\pm 6.26}$ \\ 

\hline\bottomrule
\end{tabular}
\end{center}
\caption{\textbf{Comparisons on diversity.} We compare the average perceptual divergence \minorrev{(with scaled std errors in unit of $10^{-2}$ as subscript)} of $50k$ image pairs generated by our SoFGAN vs. SOTAs. SofGAN can produce a more diverse class of styles on both CelebAMaskHD~\cite{CelebAMask-HQ} and FFHQ~\cite{karras2019style} dataset.}
\label{tb:lpips_evaluation}
\end{table*}

\begin{figure*}[ht]
\begin{center}
   \includegraphics[width=\linewidth]{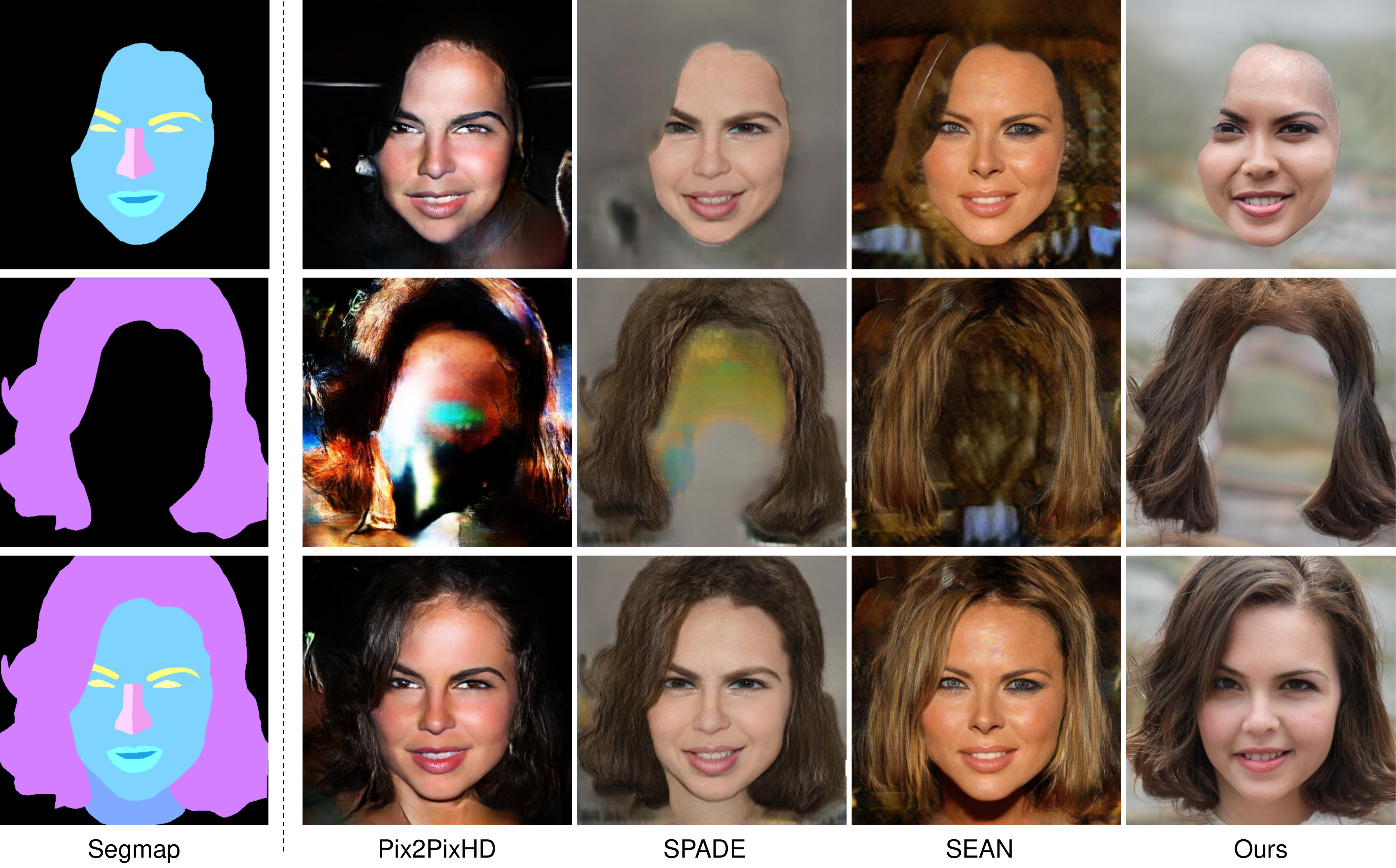}
\end{center}
   \caption{Image synthesis from partial segment maps using SofGAN vs. SOTA. The SIW-StyleGAN in our SofGAN further improves visual realism.}
\label{fig:siw_partial}
\end{figure*}

\section{Experiments}
\label{experiments}
\revised{
Next, we discuss our implementation details, conduct quantitative evaluations, and demonstrate our technique in a variety of applications.}

\subsection{Datasets}
We use the following datasets in our experiments: 

1) \textbf{CelebAMask-HQ} \cite{CelebAMask-HQ} that consists of $30k$ facial images and corresponding segmentation maps with attributes from 19 semantic part classes. We merge left/right pairs of parts into the same label (e.g., \textit{left eye} and \textit{right eye} are merged into a single \textit{eye} class) and divide the nose region into left/right parts to emphasize the geometry structure in nose region. 

2) \textbf{FFHQ} \cite{karras2019style} that consists of $70k$ high-quality images. We label the semantic classes with a pre-trained face parser. 

3) \textbf{Self-captured.} Portrait scans exhibit relatively small geometric variations after segmented into semantic regions. This enables us to use a small 3D dataset to form the sample space of the \textit{SOF}. Consequently, we capture $122$ portrait scans with manually labeled semantic properties on their texture maps (as shown in the top right of Fig. ~\ref{fig:pipeline}). We render $64$ random views for each scan to produce $7,808$ segmentation maps in total for the \textit{SOF} training. These 3D scans together with their corresponding semantic labels are ready to be released to the community.

\subsection{Implementation} 
\label{sec: implementation}

\textbf{Training the SOF.} Our \textit{SOF} consists of three trainable sub-modules: hyper-net, ray marcher, and classifier, as shown in Fig.~\ref{fig:sof_architecture}. The hyper-network $\mathcal{H}$ in Eq.~\ref{eq:hyper} contains $3$ layers with $256$ channels, and generates the weight of $\Theta$ for each $z^g$ via $4$ fully connected (FC) layers. The ray marcher takes spatial features and ray directions as input and predicts the marching \minorrev{step size}, as shown the middle of Fig.~\ref{fig:sof_architecture}. The classifier is an FC layer with $256/k$ dimensional input$/$output channels respectively, as in Eq.~\ref{eq:SOF_sep}. We use Adam optimizer~\cite{kingma2014adam}, linear warm-up and cosine decay learning rate scheduler with  $lr=1e-4$. For 122 scans, it takes about 10 hours to train \textit{SOF} till convergence (with $mIOU>0.95$).

\textbf{Training the SIW-StyleGAN.} We adopt the same settings for training as StyleGAN2~\cite{karras2019analyzing} in the following aspects: the dimensions of $z^t$, the architecture of the basis transformation network $\mathcal{W}$ (contains $8$ fully connected layers), the leaky ReLU activation with $\alpha = 0.2$, the exponential moving average of generator weights~\cite{karras2019analyzing}, the non-saturating logistic loss~\cite{goodfellow2014generative} with R1 regularization~\cite{mescheder2018training}, and the Adam optimizer with $\beta_1 = 0,\beta_2 = 0.99, \epsilon = 10^{-8}$. We refer the readers to StyleGAN2~\cite{karras2019analyzing} for more details on the training process. It is worth mentioning that we only use the discriminator to minimize the distribution distance between our SIW-StyleGAN outputs and real captured photos, which shows that the SIW Conv layer implicitly learns the alignment between segmaps and texture styles, thus \minorrev{relaxing} the pair-wise constraint between segmaps and portrait photos in training.
 
We perform all training with path regularization every $4$ steps, style mixing with $p=0.9$, and data augmentation via random scaling ($1.0 - 1.7$) and cropping. Our $1024 \times 1024$ model takes about 22 days to train on 4 RTX 2080 Ti GPUs and CUDA 10.1 ($10000kimg$ iterations \footnote{$kimg$ evaluates how many images used in the discriminator, which is independent of batchSize and GPU number}). \revised{We observe that our module is already able to obtain visually pleasing results with only $1500kimg$ iterations (by image, which takes about three days), and the rest training iterations mostly target on improving high-frequency details, such as lighting, hair, pores, etc.}

\begin{figure}[t]
\begin{center}
   \includegraphics[width=\linewidth]{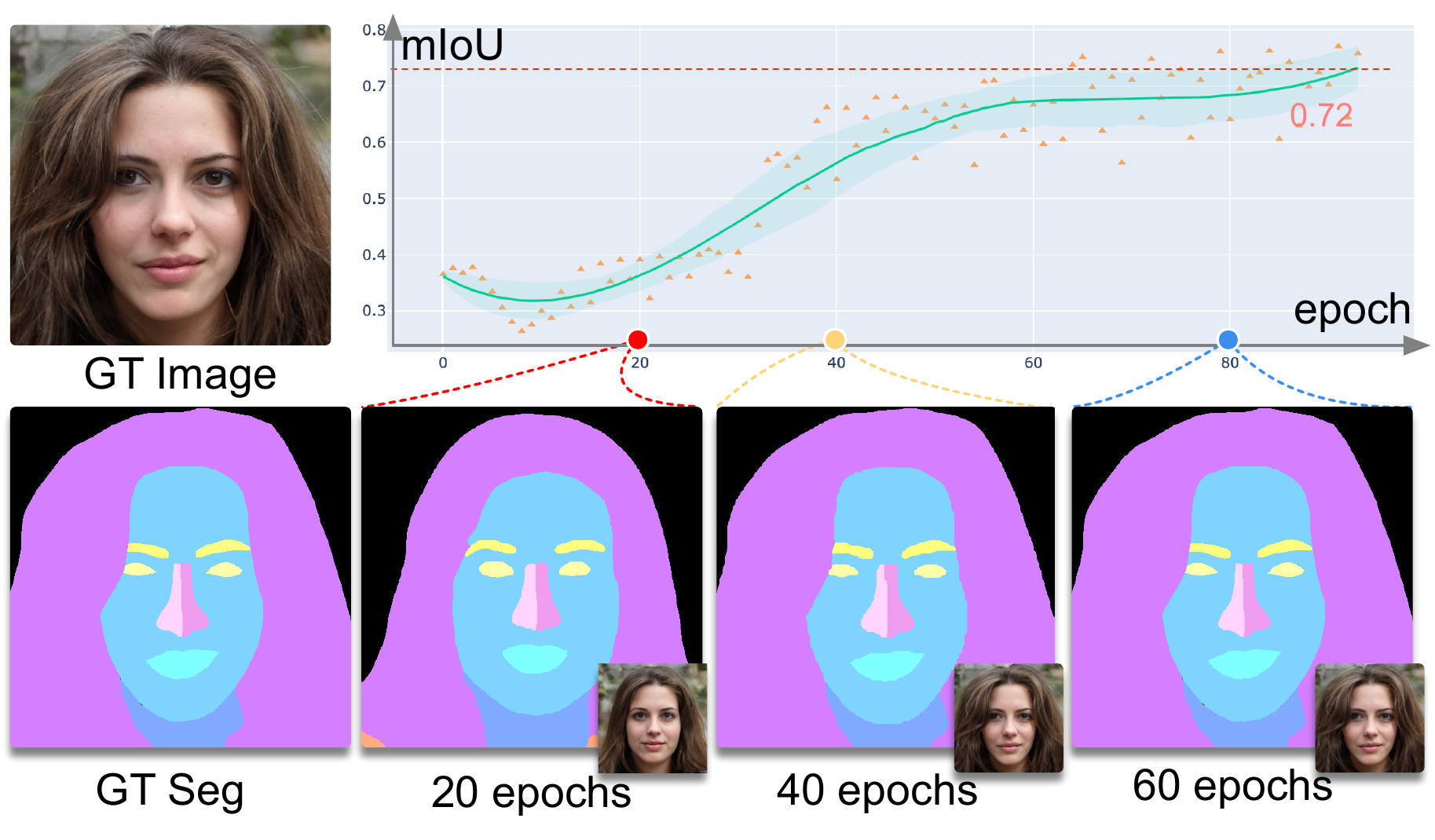}
\end{center}
   \caption{\revised{Representability evaluation of the \textit{SOF} space. We reproject 4000 segmaps from CelebAMask-HQ to our \textit{SOF} space and evaluate their $mIOU$ scores. (a) a semantic segmentation sample and its textured image using the SIW-StyleGAN. (b) Generated image and segmentation map after 20 epochs. (c) Generated image and segmentation map after 40 epochs ($mIoU>0.5$). (d) Final regressed segmentation map and generated image.}}
\label{fig:sof_miou}
\end{figure}

\begin{figure*}[t]
\begin{center}
   \includegraphics[width=\linewidth]{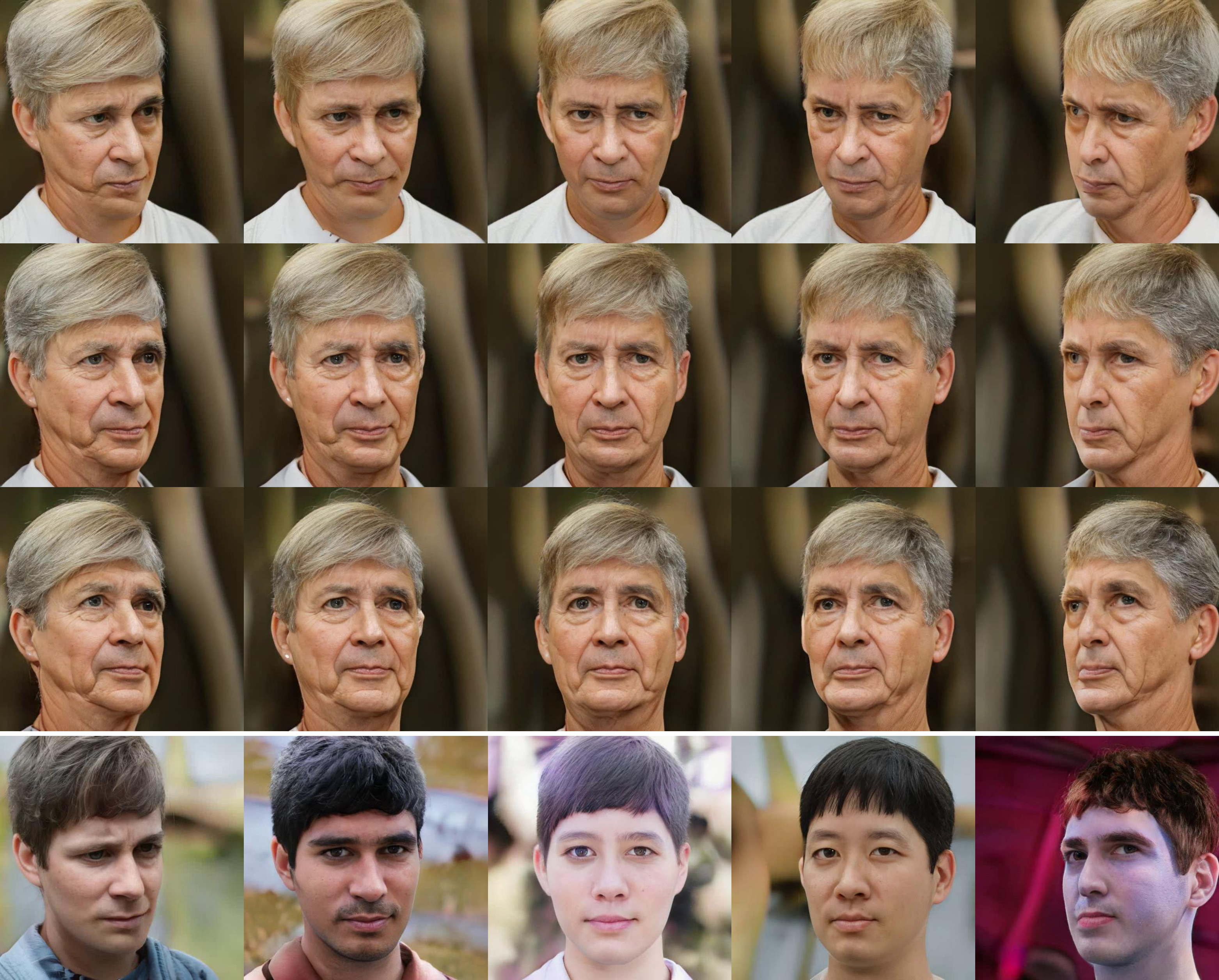}
\end{center}
   \caption{\revised{Free-viewpoint generation results on a same \textit{SOF}. The first three rows show the free-viewpoint results with a same texture style and the last row shows results of applying different texture styles to the \textit{SOF}. Our free-viewpoint stylizing effects conform the perspective imaging rules and also is able to preserve the facial identity. }}
\label{fig:SDFStyles}
\end{figure*}

\begin{figure*}[t]
\begin{center}
   \includegraphics[width=\linewidth]{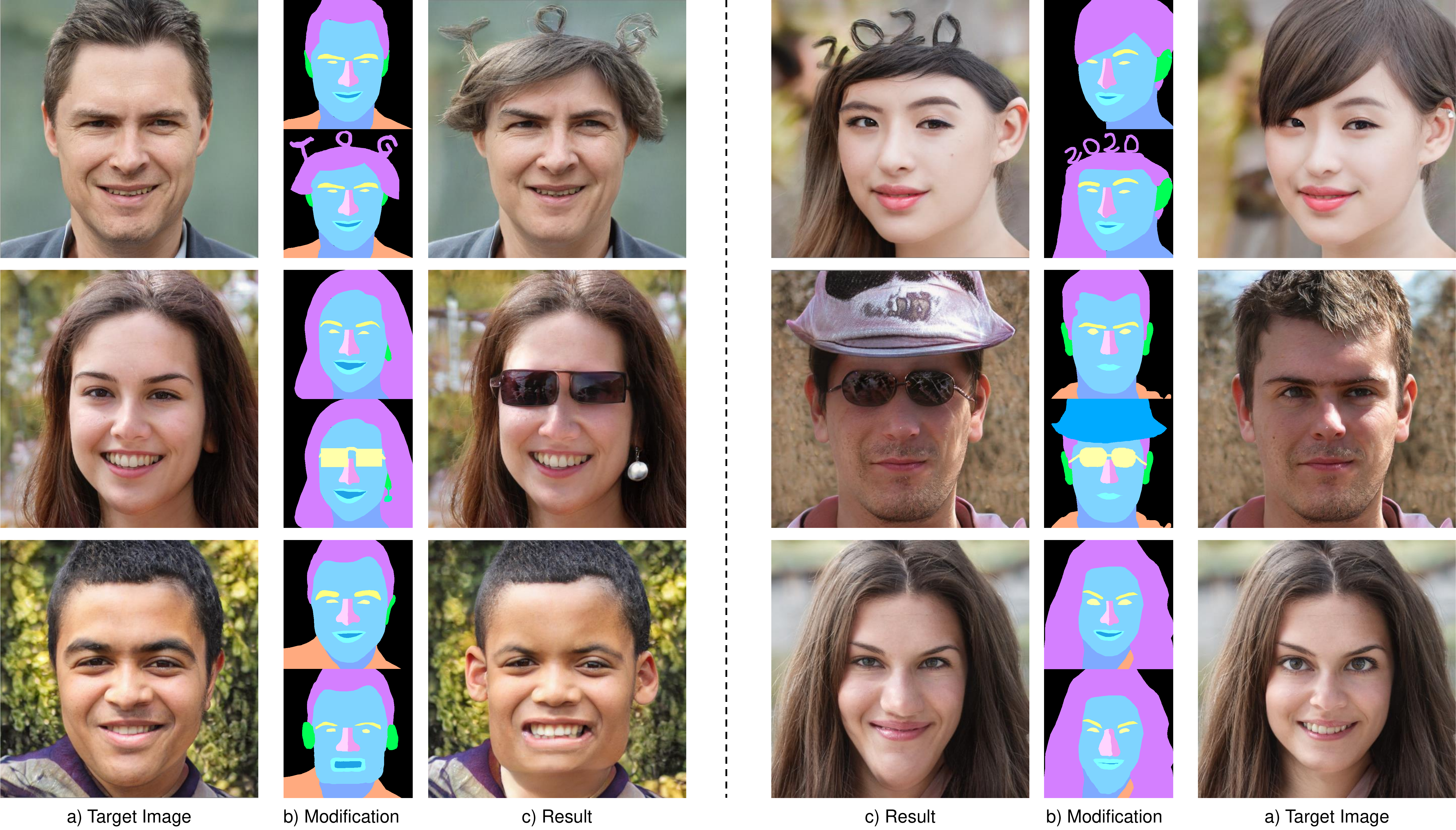}
\end{center}
   \caption{\revised{Interactive image generation. Our method enables interactively synthesizing photo-realistic portrait images by drawing semantic segmentation maps (b columns). Our region-based texturing scheme is able to preserve shape and generalizes well even for extremely unnatural segmaps (as the "\textit{TOG}" and "\textit{2020}" patterns in the first row). In the second row, we add glasses and an earring to the left figure while we add glasses and a hat to the right figure. We can observe that our method can dynamically adjust the Appearance (even the lighting) according to the manipulation. In the last row, we manipulate the facial shapes and expressions.}}
\label{fig:drawing}
\end{figure*} 

\begin{figure*}[t]
\begin{center}
   \includegraphics[width=\linewidth]{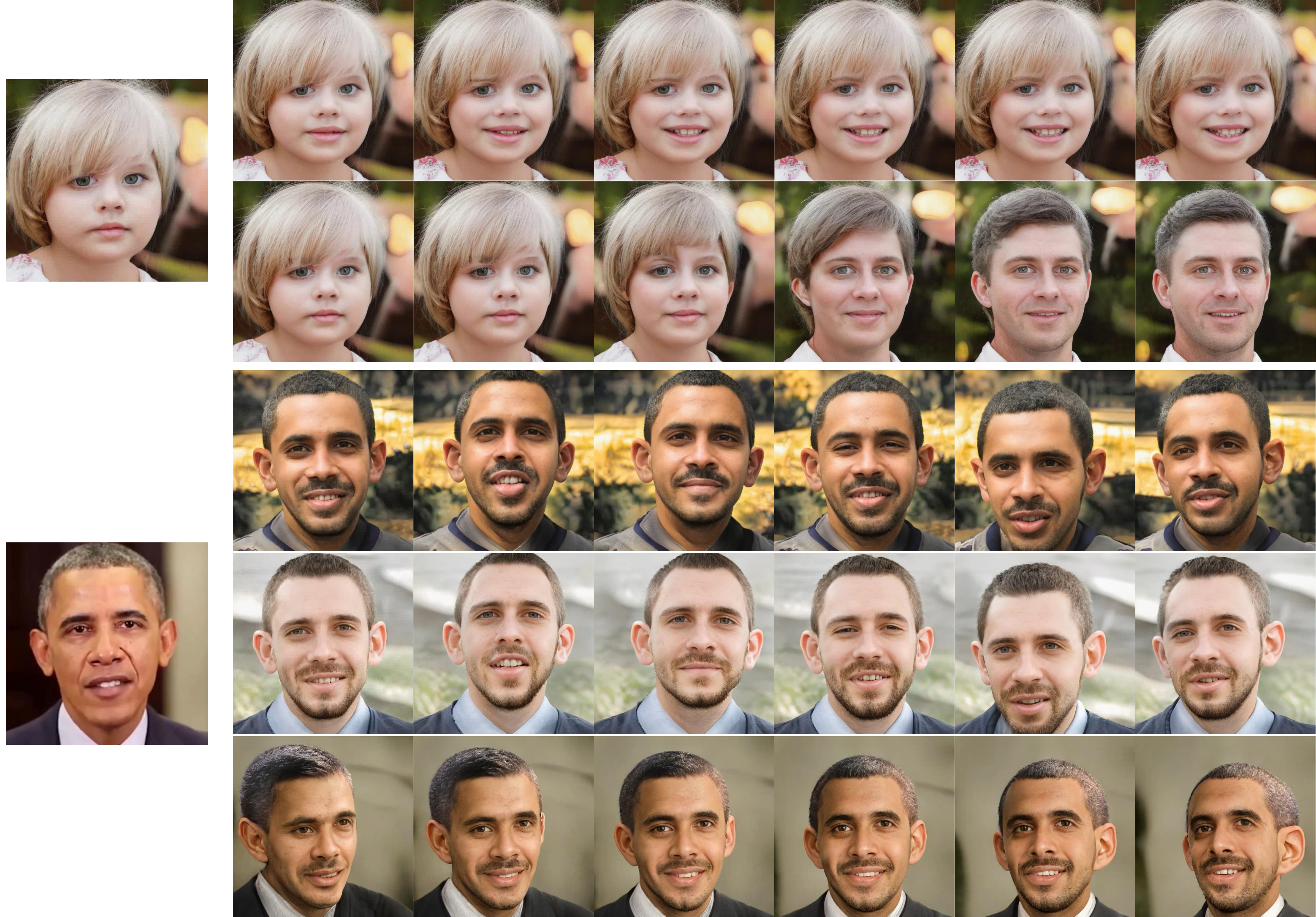}
\end{center}
   \caption{Our method preserves geometric and textural consistency in several applications. In expression editing (1st row), the hairstyle and background remain unchanged when the character changes from a neutral expression to a smiling). In gender/age morphing (2nd row), we maintain photo-realism of the central shape. In style mixing (3rd and 4th rows), we adopt two different styles and our results maintain consistency across frames (see the supplemental materials for comparison between ours vs. SOTA). In free-viewpoint rendering (bottom row), we first parse the segmap of President Obama's photo, then back project the results to the \textit{SOF} latent space, and finally stylize the segmaps with a random texture style.}
\label{fig:attributes}
\end{figure*} 

\subsection{Quantitative Evaluation.}
\label{sec: quantitative}

We quantitatively evaluate our method on various metrics including the Fréchet Inception Distance (FID)~\cite{heusel2017fid}, Learned Perceptual Image Patch Similarity (LPISP) ~\cite{zhang2018perceptual}, mIOUs and facial identity metrics. We compare our method with several recent image synthesis methods: Pix2PixHD~\cite{wang2018pix2pixHD}, SPADE~\cite{park2019SPADE}, SEAN~\cite{zhu2020sean}, and the baseline StyleGAN2~\cite{karras2019analyzing}. 

Table~\ref{tb:comp_input} shows the requirements and achievable effects of each method. In contrast to the competing methods, our SIW-StyleGAN enables both global and local style adjustments without requiring \minorrev{pairwise} data for training. However, we require an additional small dataset with 3D segmaps to form the basis of our geometric generation space.

To evaluate the above methods under the same settings of dataset, iterations, and resolution, we retrain their modules on the FFHQ~\cite{karras2019style} and CelebA~\cite{karras2017progressive} datasets with $800kimg$ iterations at $512^2$ resolution, and swap their segmentation maps as the evaluation set (i.e., if trained with dataset A, then use the segmentation maps of dataset B for testing). We set $truncation=1.0$ and calculate FID value on $50k$ randomly sampled images. As shown in Fig.~\ref{fig:curve}, our method achieves the best FID scores on both FFHQ and CelebA datasets\footnote{The results for the comparison methods might be slightly different from their original papers, as we use random styles during evaluation.}. 

We attribute our major improvements to: 
\begin{enumerate}
    \item Different from SPADE and SEAN, our SIW-StyleGAN is trained with a non-pair-wise and unsupervised scheme instead of a perception loss on the pair-wise target photos, resulting in more freedom in the texture style space and thus increasing the variety of the generated images.
    \item Since we divide the image generation into semantic regions which have relatively similar texture patterns (i.e., semantic regions with the same class share a similar data distribution), our generator is able to focus on each small region separately instead of handling the whole image at the same time, and therefore, it converges faster than StyleGAN2.
\end{enumerate}

\revised{
Fig.~\ref{fig:siw_partial}, Fig.~\ref{fig:comparision_celeba} and Fig.~\ref{fig:comparision_ffhq} in the Appendix give several visual comparisons of our method v.s. recent conditional image synthesis methods: Pix2PixHD~\cite{wang2018pix2pixHD}, SPADE~\cite{park2019SPADE}, SEAN~\cite{zhu2020sean}\footnote{SEAN extracts style codes from reference style images, while other methods use randomly sampled style codes as input. Thus, we project the images in the evaluation set to the texture sampling space and acquire style codes for each image. To satisfy the pairwise constraint on semantic classes in SEAN, we only use the style images that have the same semantic classes with the conditioning semantic maps for the generation.}, and StyleGAN2~\cite{karras2019analyzing} on both CelebAMask-HQ~\cite{CelebAMask-HQ} and FFHQ~\cite{karras2019style} datasets. From the comparisons, we can see that our framework performs better than Pix2PiXHD, SPADE, and SEAN in both style diversity and realism of the generated images with comparable quality as StyleGAN2, which only supports global stylization.}

\revised{
Besides the improved realism, SofGAN further improves the diversity of texture styles. We generally expect the GANs to produce images with a wide variety while training GANs, a commonly encountered issue is that the discriminator gets stuck in a local minimum and the generator starts to produce similar outputs to favor the specific discriminator, i.e., the \textit{Mode Collapse} problem. During the experiment, we found that decomposing the generation space into sub-spaces and regionally generation could also help for dealing with the \textit{Mode Collapse} problem and improve diversity. }

\revised{
To evaluate the representability of the trained SOF space, we therefore first randomly select $4,000$ segmap samples from the CelebAMask-HQ dataset \cite{CelebAMask-HQ} and then search for the corresponding geometry latent code in our geometric sampling space with the mIOU metric for similarity evaluation. Fig. ~\ref{fig:sof_miou} shows that, for most samples, our scheme manages to find the corresponding geometry code with around $6k$ iterations. This indicates that the space formed with $122$ 3D scans can already cover a large variety of portrait shapes. We then jointly optimize $\Phi$ and $\Theta$ by casting rays from randomly sampled views for each instance $D_i$ in $\mathcal{D}$ and calculate the cross-entropy loss between the predicted segmentation and the ground truth segmentation rendered from $D_i$. }

\revised{
To evaluate the diversity among the generated images, we randomly sample $50k$ image pairs\footnote{With 50k image pairs we can constraint the standard derivative of LPIPS for all pairs at around $+/- 0.0003$, which we think is small enough to represent the general statics of the whole generation space.} from the same checkpoint with the FID evaluation ($800kimg$ iterations), and calculate the mean LPIPS~\cite{zhang2018perceptual} value with three backbone architectures (VGGG16, AlexNet, and SqueezeNet). Table ~\ref{tb:lpips_evaluation} compares our result with several recent GAN-based image generators. A higher LPIPS value means the generated images are more diverse, while a lower score means more similar to each other. \minorrev{Considering that} artifacts and image noise could also pull up the LPIPS score, we compare the FID and LPIPS score simultaneously, and observe that our method achieves better image quality (FID score) with more diversity (\minorrev{LPIPS} score).}

\revised{
Next, we evaluate the diversity and identity of conditional image generation by controlling the segmap and poses. For the diversity evaluation, we first randomly sample $1k$ segmaps from CelebaMask dataset and stylize the above each random sampled segmap with $6$ random texture styles and calculate the LPIPS ~\cite{zhang2018perceptual} metric between the generated images ($10$ random image pairs for each). We show qualitative and quantitative results in Fig. ~\ref{fig:ssg_comp} and the  average LPIPS scores for single segmap generation (i.e., the SSG Diversity) of Table ~\ref{tb:fvv_eval}. Fig. ~\ref{fig:ssg_comp} shows that our SofGAN can generate more realistic and less artifact results.}

\revised{
To evaluate the view consistency metric among the generated free viewpoint images, we randomly sample $1k$ shape instances from the \textit{SOF} spaces and render $15$ views pure instance (e.g.. as shown in the first three rows of Fig. \ref{fig:SDFStyles}). Then, we randomly sample $10$ image pairs of the $15$ views and evaluate their similarity score with the dlib face recognition algorithm \footnote{\url{https://github.com/ageitgey/face_recognition}}. Note that we center crop face region with a face detection scheme before the comparison to favorite the input setting of the face parser. Table. \ref{tb:fvv_eval} gives the quantitative evaluation for the free viewpoint video generation,. We observe that our SIW module is able to better preserve the facial identity compared with the existing method \cite{zhu2020sean,park2019SPADE}. }

\definecolor{Blue}{rgb}{0.00,0.00,1.00}
\begin{table}[ht]
\begin{center}
\begin{tabular}{c|cc|cc}
\toprule\hline
& \multicolumn{2}{c|}{\textbf{FVV Identity $\uparrow$}} & \multicolumn{2}{c}{\textbf{SSG Diversity $\uparrow$}}\\ \cline{2-5}
& CelebA & FFHQ & CelebA & FFHQ \\ \hline
\textbf{SPADE}& $0.460_{\pm 12.3}$ & $0.414_{\pm 10.8}$ & $0.116_{\pm 4.87}$ & $0.085_{\pm 7.41}$ \\ 

\textbf{SEAN}  & $0.438_{\pm 11.8}$ & $0.399_{\pm 10.1}$ & $0.463_{\pm 0.18}$ & $\textbf{0.534}_{\pm 0.14}$  \\ 

\textbf{SofGAN}   & $\textbf{0.471}_{\pm 13.3}$ & $\textbf{0.448}_{\pm 12.8}$ & $\textbf{0.463}_{\pm 6.17}$ & $0.480_{\pm 5.42}$ \\ 
\hline\bottomrule
\end{tabular}
\end{center}
\caption{\textbf{Quantitative evaluation of controlled generation.} We compare the preservation of facial identity under free viewpoint video generation (FVV Identity), and style variance generated with single segmap generation (SSG Diversity) with SPACE\cite{park2019SPADE} and SEAN\cite{zhu2020sean}. Our architecture is able to better preserve identity during free-viewpoint generation and achieve compatible image diversity with SEAN when conditional on a same segmap. \minorrev{We show the mean FFV/SSG values of 1k images in each cell with std error as subscript in the unit of $10^{-2}$.}}
\label{tb:fvv_eval}
\end{table}

\begin{figure*}[t]
\begin{center}
   \includegraphics[width=\linewidth]{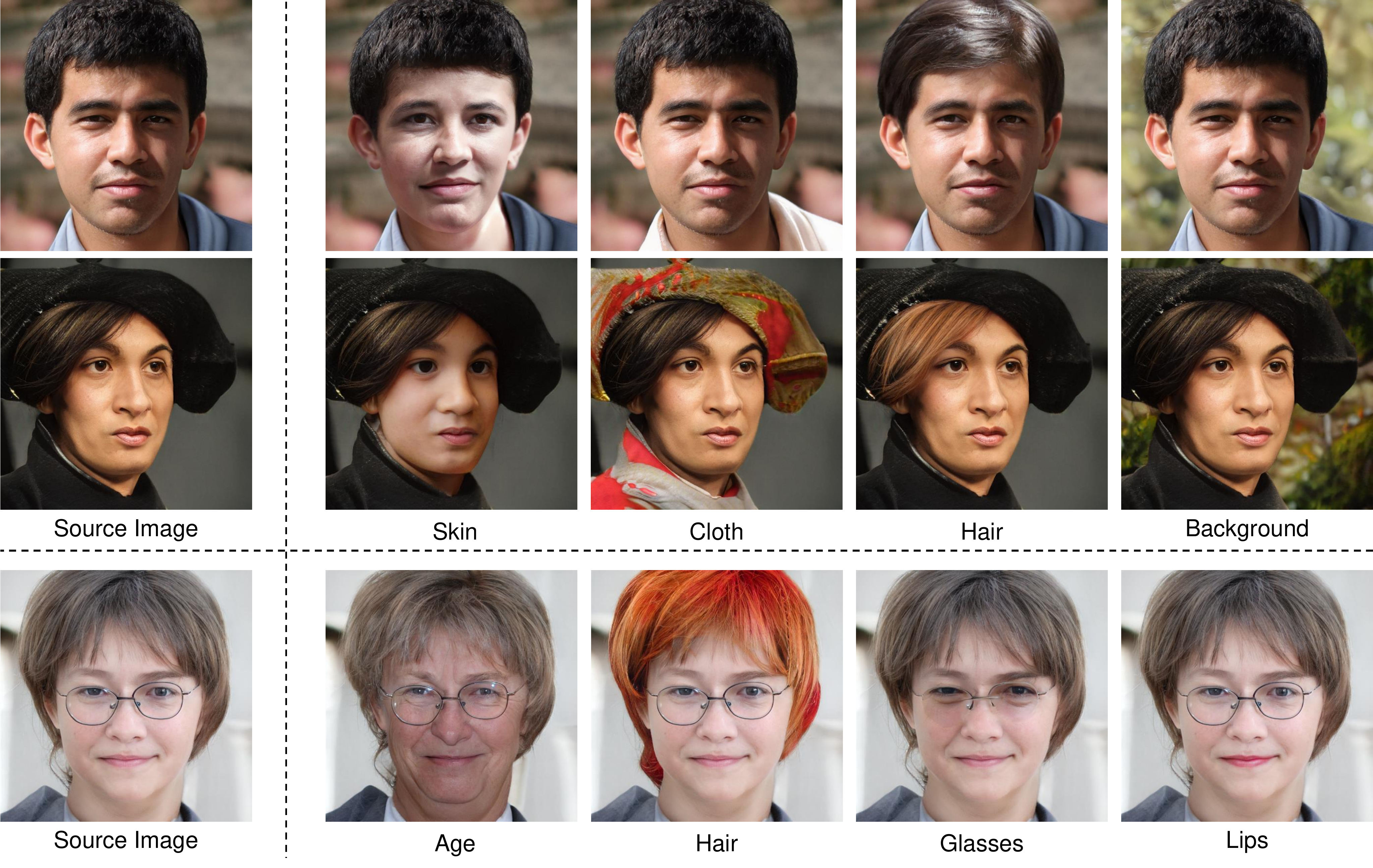}
\end{center}
   \caption{Semantic-level style adjustment. Our algorithm enables semantic level style adjustment on $17$ classes: skin, hair, lips, eyes \& eyebrows, wearings, background and etc..}
\label{fig:application on texture space}
\end{figure*} 

\subsection{Applications}
\label{sec:applications}

Recall that our generator is controlled by three individual components: camera poses $C[R, T, K]$, shape space latent code $z^g$, and texture space latent code $z^t$. In the following section, we will demonstrate various applications by exploring the above three components.
 
\textbf{Free-Viewpoint Video Synthesis.} Under our generation framework, we can generate free-viewpoint portrait images from geometric samples or real images by changing the camera pose.

As shown in Fig.~\ref{fig:SDFStyles}, the last row of Fig.~\ref{fig:attributes} and Fig.~\ref{fig:free-viewpoint}, since our \textit{SOF} is trained with multi-view semantic segmentation maps, the geometric projection constraint between views is encoded in the \textit{SOF}, which enables our method to keep shape and expression consistent when changing the viewpoint. \revised{Besides, our model is able to preserve overall consistency (regardless of some local surface details) of texture style including facial appearance and hairstyle even under significant view angle variation. }
\revised{
For real images, we first parse a monocular segmap from the image, and reproject the segmap back to the geometric space. Then we generate SOF for the segmap and render free-view segmaps. The last row of Fig.~\ref{fig:attributes} shows an free-viewpoint Obama result and Fig.~\ref{fig:seg_mv} gives several examples for photos from CelebA dataset and internet.}

\textbf{Shape Space Exploration.} 
\label{sec:Shape generation and interpolation}
As shown in Fig. ~\ref{fig:attributes} and Fig. ~\ref{fig:drawing}, our method also supports generation and attribute editing in both 2D image space and the 3D geometric space. To generate new shapes, we fit a Bayesian Gaussian Mixture Model (GMM) with Dirichlet Process Prior to all samples $z^{g}$ in the trained dispersed geometric sampling space $\mathcal{G}$, and then sample from the continuous GMM to generate new shapes outside the training dataset. For attribute editing, we borrow the idea of attribute decoupling~\cite{shen2019interpreting} and extract shape-related eigenvectors (e.g., expression, age) from the \textit{SOF} $z^{g}\in\mathbb{R}^{256}$ via principal component analysis. The editing process is contiguous and allows us, for example, in the 1st row of Fig.~\ref{fig:attributes}, to gradually turn the expression of the little girl from neutral to smile.

The regional-specific property of our \minorrev{method} also helps in dealing with the entanglement issue in most existing decoupling-based methods for the controllable generation (~\cite{shen2019interpreting, shen2020interfacegan, Deng2020DisentangledAC}).  The first row of Fig.~\ref{fig:attributes} shows a smile expression interpolation example, with the general knowledge that smiling is mostly related to facial regions (e.g., eye, eyebrow and mouse regions) except other regions (e.g., hair, cloth, etc.), thus we only change the related region while fixing none-related region that enables to better preserve geometric and textural consistency during the generation. As our SIW-StyleGAN is region-aware, when applying shape morphing, the generated intermediate shape between two random shapes is also photorealistic (the 2nd row of Fig.~\ref{fig:attributes}).

Apart from sampling shapes from the \textit{SOF} latent space, our method can also generate portrait images from existing photos and videos or hand-painted 2D segmentation maps, as demonstrated in Fig.~\ref{fig:drawing} and the bottom rows of Fig.~\ref{fig:attributes}. In Fig.~\ref{fig:drawing}, we interactively modify segmentation maps to obtain novel user-specified shapes. Such an effect can benefit artistic creation or special effect generation. In Fig.~\ref{fig:attributes}, we collect a video clip from Internet and generate segmentation maps for each frame with a pre-trained face parser~\cite{yu2018bisenet}. Our method can preserve texture style and shape consistency on various poses and expressions without any temporal regularization. We refer  readers to our supplementary video for the animated sequences.

\textbf{Texture Space Exploration.} 
\label{Global and Local Style Adjustment}
One of the key features of our SIW-StyleGAN is semantic-level style controlling. Benefiting from the StyleConv blocks and style mixing training strategy, we could separately control the style for each semantic region by adjusting the composed $z^t$ through the similarity map $\mathcal{P}$. Fig.~\ref{fig:application on texture space}, ~\ref{fig:real_editing} demonstrates the dynamic styling effects by changing texture styles in background, skin, hair, facial-region, hair, lip and wearing (e.g., cloth, hat and glasses etc.) regions of source images (more results could be found in the Appendix). Still, our approach could keep the texture styles unchanged except for the target region and adaptively adjust the semantic boundaries to ensure that the output looks natural. Moreover, our method could preserve global lighting when adjusting regional styles. Please refer to our supplemental video for more effects.

To better demonstrate the performance of our method, we further train a SIW-StyleGAN model with $10000kimg$ iterations (one image per batch) under $1024^2$ resolution. Fig.~\ref{fig:global_style_1}, \ref{fig:global_style_2} shows more results on global style adjustment, Fig.~\ref{fig:regional_styles} shows regional style controlling during generation, while Fig.~ \ref{fig:real_editing} shows real-captured photo editing via our method.

\begin{figure}[ht]
\begin{center}
  \includegraphics[width=\linewidth]{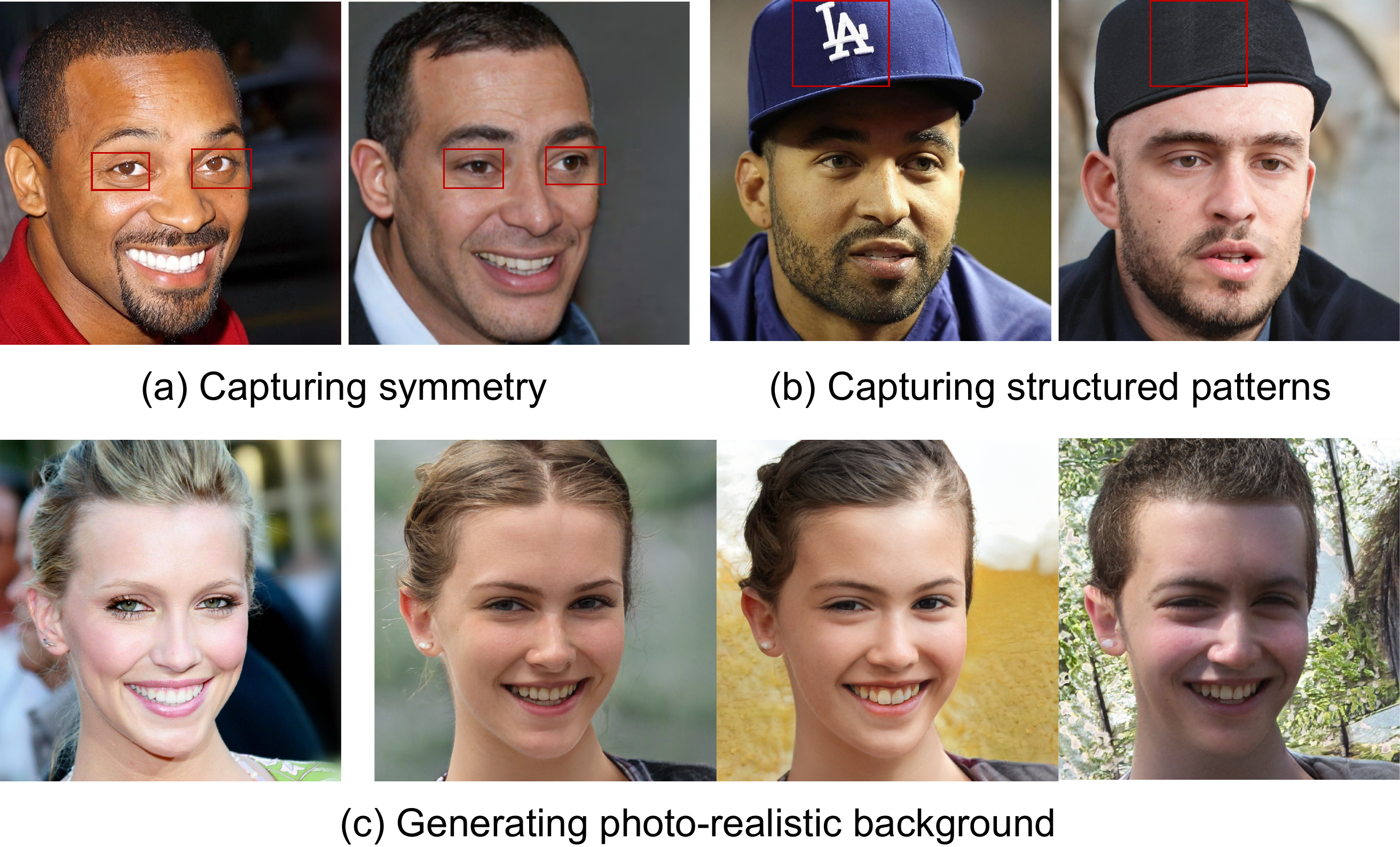}
\end{center}
\caption{\revised{We explore the limited power of SofGAN: (a) Capturing symmetry. The eyes may look towards different directions when generating extreme side faces(this phenomenon can be also seen in the right side of Fig. ~\ref{fig:SDFStyles}). (b) Capturing well-structured patterns. The SofFGAN tends to generate flatten textures in decorative regions, and failed to reconstruct structured patterns like the \textit{"LA"} in the hat. (c) Generating photo-realistic background, compared with real-captured photo (left), the generated background region tends to be blurry or noisy. Note that the left sides of each sub-figure are real-captured photos while the right sides are our generated images.}}
\label{fig:limitations}
\end{figure}

\section{Conclusion}

In this paper, we have presented a novel two-stage portrait image generation framework that enables dynamic styling. We employed a semantic occupancy field as \minorrev{the geometric} representation and a semantic \minorrev{instancewise} StyleGAN for regional texturing. With our decomposed generation framework, we not only enable attribute-specific control over the generation but also allow users to generate from existing segmaps and editing attributes of existing images. In particular, our approach implicitly learns 3D geometric priors from 2D semantic maps without 3D supervision, eliminating the need for high-quality 3D portrait scans. Besides, our unsupervised training scheme for SIW-StyleGAN does not require paired segmaps and photos for training. Comprehensive experiments have shown that our approach achieves SOTA FID and LPISP scores on both CelebA and FFHQ datasets and can be used in a broad class of synthesis tasks. 

By using disentangled geometric and texture space, we could guarantee multi-view consistency at geometry level with \textit{SOF}. However, as SIW-StyleGAN texture each semantic region independently, we do not guarantee pixel-level multi-view consistency, \minorrev{ and significant shape modification would also leads to noticeable texture changes (Fig. ~\ref{fig:siw_partial}, last column).} Moreover, our method still exhibits limitations in capturing symmetry, structured patterns and photo-realistic generation in the background region. For example, the gazing directions sometimes are inconsistency (e.g., Fig. ~\ref{fig:limitations} (a)). Also, our method tends to output either flattened or regionally repetitive texture in the same semantic region, thus perform poorly on synthesizing complex patterns within the same semantic region, like the \textit{"LA" pattern in Fig. \ref{fig:limitations} (b). As shown in Fig. \ref{fig:limitations} (c), though our method succeeds in generating photo-realistic textures for the facial region, the background region is generally noisy or blurry.} This may originate from the fact that our discriminator is not region-aware and only discriminates at the global distribution. As a result, such a design would flatten semantically specified styles. We plan to \minorrev{redesign} the discriminator to be region-based and improve the semantically specified structures and styles in future work.

\section{Acknowledgements}
We thank Xinwei Li, Qiuyue Wang for dubbing the video, Zhixin Piao for comments and discussions, as well as Kim Seonghyeon and Adam Geitgey for sharing their StyleGAN2 implementation and face recognition code for our comparisons and quantity evaluation. This work was supported by NSFC programs (61976138, 61977047), the National Key Research and Development Program (2018YFB2100500), STCSM (2015F0203-000-06) and SHMEC (2019-01-07-00-01-E00003).

\bibliographystyle{ACM-Reference-Format}
\bibliography{sample-base}

\appendix

\section{Network Architectures}
\revised{Table ~\ref{tbl:net_sof} shows detailed specifications of each sub-module in the neural \textit{SOF} representation described in Sec.~\ref{SOF Network Approximate} of the main manuscript. Table ~\ref{tbl:net_siw} shows detailed network specifications for the texture modeling described in Sec.~\ref{sec:siw stylegan} of the main manuscript.}

\definecolor{Blue}{rgb}{0.00,0.00,1.00}
\begin{table}[ht]
\begin{tabular}{cccc}
\toprule\hline
& \textbf{Layer}  & \textbf{Channels} & \textbf{Input} \\ \hline
\multirow{3}{*}{\textbf{T}} 
  & $fc_{0\_0}$   & 3/256  & point location $x$ \\
  & $fc_{0\_1}$   & 256/256  & $fc_{0\_0}$ \\ 
  $\Theta$ & $fc_{0\_2}$   & 256/256  & $fc_{0\_1}$ \\ \hline
\multirow{5}{*}{\textbf{\textit{RayMarcher}}} 
  & $fc_{1\_0\_l}$   & 256/256   & $fc_{0\_2}$ \\
  & $fc_{1\_0\_b}$     & 3/256   & ray\_dir $\vec{d}$ \\
  & $fc_{1\_i\_l}$   & 256/256   & $fc_{1\_{(i-1)}\_l}+fc_{1\_{(i-1)}\_b}$ \\
  $i \in [1,5]$ & $fc_{1\_i\_b}$     & 3/256   & ray\_dir $\vec{d}$ \\
  & $fc_{1\_6\_l}$   & 256/1     & $fc_{1\_5\_l}+fc_{1\_5\_b}$ \\
  & $fc_{1\_6\_b}$     & 3/1     & ray\_dir $\vec{d}$ \\
  &&& $fc_{1\_6\_l} + fc_{1\_6\_b}$\\\hline
\multirow{3}{*}{\textbf{\textit{Classifier}}} 
  & $fc_{2\_0}$    & 256/256    & $fc_{0\_2}$ \\
  & $fc_{2\_1}$    & 256/256    & $fc_{2\_0}$ \\
 $\Phi$  & $fc_{2\_2}$    & 256/256    & $fc_{2\_1}$ \\
  & $fc_{2\_3}$   & 256/20     & $fc_{2\_2}$ \\
  &&&$fc_{2\_3}$\\\hline

\bottomrule
\end{tabular}
\caption{\revised{\textbf{Layer specifications for the neural \textit{SOF} representation.} \textit{fc} denotes a fully-connected layer with \textbf{\textit{Layer Normalization}} and \textbf{\textit{ReLU}} activation.}}
\label{tbl:net_sof}
\end{table}

\begin{table}[ht]
\begin{tabular}{ccccc}
\toprule\hline

& \textbf{Layer} & \textbf{Channels}& \textbf{s} & \textbf{Input} \\ \hline
\multirow{15}{*}{\textbf{\textit{SIW-StyleGAN}}} 
  & $Conv_{0\_0}$ &  16 & 2 & $\mathcal{M}$\\
  & $Conv_{0\_1}$ &   16 &  1  & $Conv_{0\_0}$\\
$i \in [1,3]$ & $Conv_{i\_0}$ &  $16\times2^i$ & 2 & $Conv_{(i-1)\_0}$\\
  & $Conv_{i\_1}$ &   $16\times2^i$ &  1  & $Conv_{i\_0}$\\
  
    & $\Psi_0$ &  512 &  $1/2$ & $z^t, Conv_{3\_1}$\\
$j \in [1,3]$    & $\Psi_j$ &  512 &  $1/2$ & $z^t, \Psi_{j-1}$\\
& $\Psi_4, \mathcal{N}$ &  512 &  $1/2$ & $z^t, \Psi_3, \mathcal{M}, \mathcal{P}$\\
& $\Psi_5, \mathcal{N}$ &  256 &  $1/2$ & $z^t, \Psi_4, \mathcal{M}, \mathcal{P}$\\
& $\Psi_6, \mathcal{N}$ &  128 &  $1/2$ & $z^t, \Psi_5, \mathcal{M}, \mathcal{P}$\\
& $\Psi_7$ &  64 &  $1/2$ & $z^t, \Psi_6, \mathcal{P}$\\
& $\Psi_8$ &  32 &  $1/2$ & $z^t, \Psi_7, \mathcal{P}$\\
& $\Psi_9$ &  3 &  $1/2$ & $z^t, \Psi_8, \mathcal{P}$\\
\hline
\bottomrule
\end{tabular}

\caption{\revised{\textbf{Network architecture for texture modeling (SIW-StyleGAN).} \textbf{s} denotes stride size and $\mathcal{M}$ denotes the semantic segmentation map. $z^t \in \mathcal{R}^{2\times512}$ and $\Psi$ is the texture style modulation function as mentioned in Sec.~\ref{sec:siw stylegan}. $\mathcal{N}$ is the spatially adaptive normalization layer while $\mathcal{P}$ is the style distance map for the mixed style training.}}
\label{tbl:net_siw}
\end{table}

\section{Ablation Study}
To better analyze the contribution of each component in our SofGAN, we conduct ablation studies on:

1) ray marching with LSTM vs. our proposed architecture.

2) constant input vs. with semantic map encoder.

3) with vs. w/o SIW style mixing block. 

The modules in these ablation studies are \textbf{only} trained with $800kimg$ iterations and $512^2$ resolution as it is challenging to train each module with $10000kimg$ and $1024^2$ due to limited computing resources. Hence the image quality demonstrated in this section is much lower than the application section (Sec.~\ref{sec:applications}).

\begin{figure*}[t]
\begin{center}
  \includegraphics[width=1.0\linewidth]{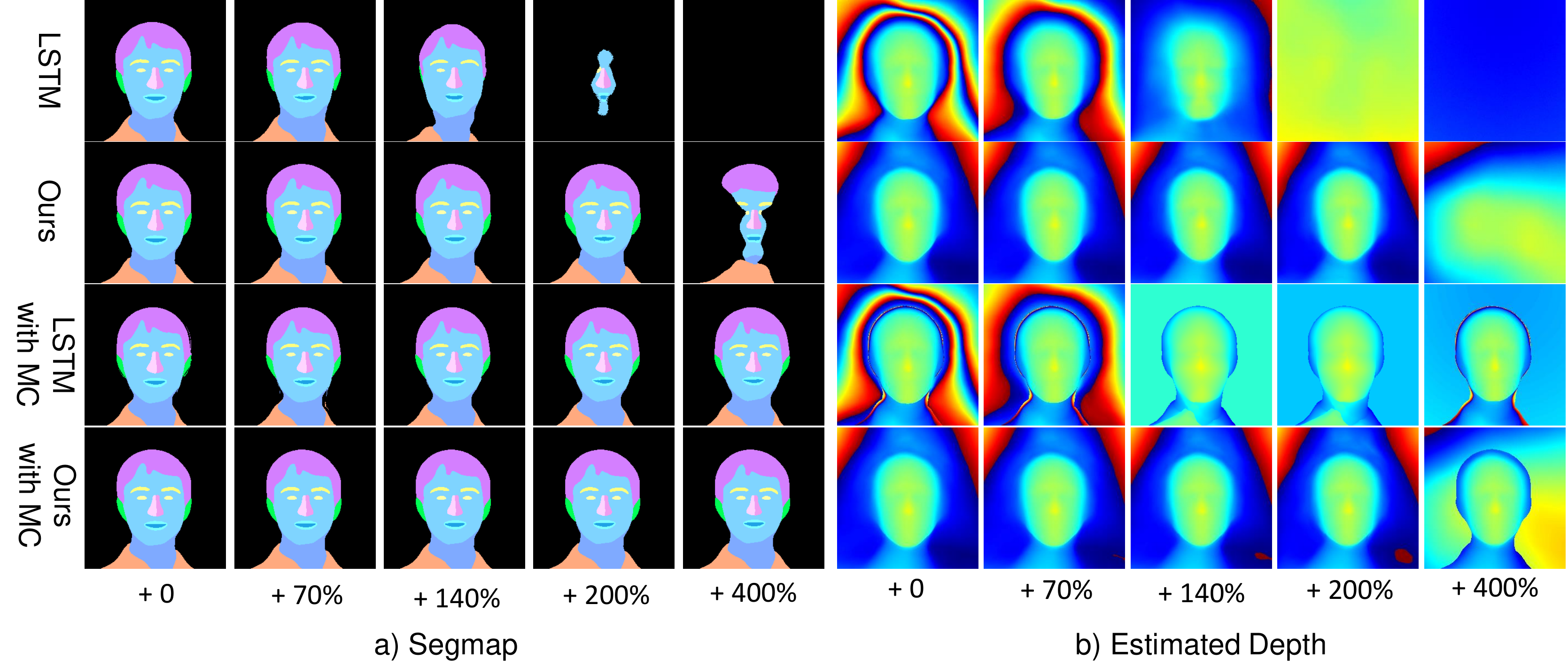}
\end{center}
  \caption{\revised{Ablation analysis on the proposed ray marching network.} It shows segmentation maps (left 5 columns) and depth maps (right 5 columns) estimated using different ray marching schemes. We gradually increase the distance rate of view camera, i.e., (novel view - training sets) / (training sets - world center). Each column shows the corresponding results under a distance rate.}
\label{fig:ablation_sof}
\end{figure*}

\textbf{Ray Marching Architecture}
\label{sec:ray_matching}

We observed that the LSTM ray marcher \cite{Sitzmann2019SceneRN} is sensitive to the initial camera position and can easily fail to render novel views when only trained \minorrev{on} sparse views. Thus, we propose a new ray marching architecture, which estimates the step size for marching only based on the current position feature and ray direction without temporal status. This section conducts an ablation study on the ray marcher by feeding with various camera positions.

As demonstrated in Fig. ~\ref{fig:ablation_sof}, we first compare LSTM approximation vs. our proposed architecture in the first two rows. The first column of Fig. ~\ref{fig:ablation_sof} (a) shows a novel camera that is close to the training cameras. For the (a) column of Fig. \ref{fig:ablation_sof} , we move the camera away from the training camera positions and decrease the Field of View (FOV) synchronously to ensure the face is roughly at the same location in the image plane. Fig. ~\ref{fig:ablation_sof} (b) shows the estimated depth $t$ for each camera position in Fig. ~\ref{fig:ablation_sof} (a). We can observe that our proposed architecture is much less sensitive to the camera position. 

The last two rows of Fig. ~\ref{fig:ablation_sof} further analyze the results when giving good depth initials, i.e., we initialize $t$ for each ray with the Marching Cube (MC) Algorithm (as shown in Fig. \ref{fig:vis_sof}). In practically, we first uniformly sample the ``background'' probability and to obtain the portrait surface via $0.5$ probability threshold, and then project the surface to the rendering camera to obtain the initial ray marching depth $t$. As shown in the last two rows of Fig. ~\ref{fig:ablation_sof} (b), LSTM can obtain fine predictions inside the object but performs poorly on the boundaries between the portrait and the background (the boundaries are eroded in this case due to wrong depth estimation). On the contrary, our architecture is able to completely recover both interior and boundary regions.

\begin{figure*}[t]
\begin{center}
  \includegraphics[width=0.83\linewidth]{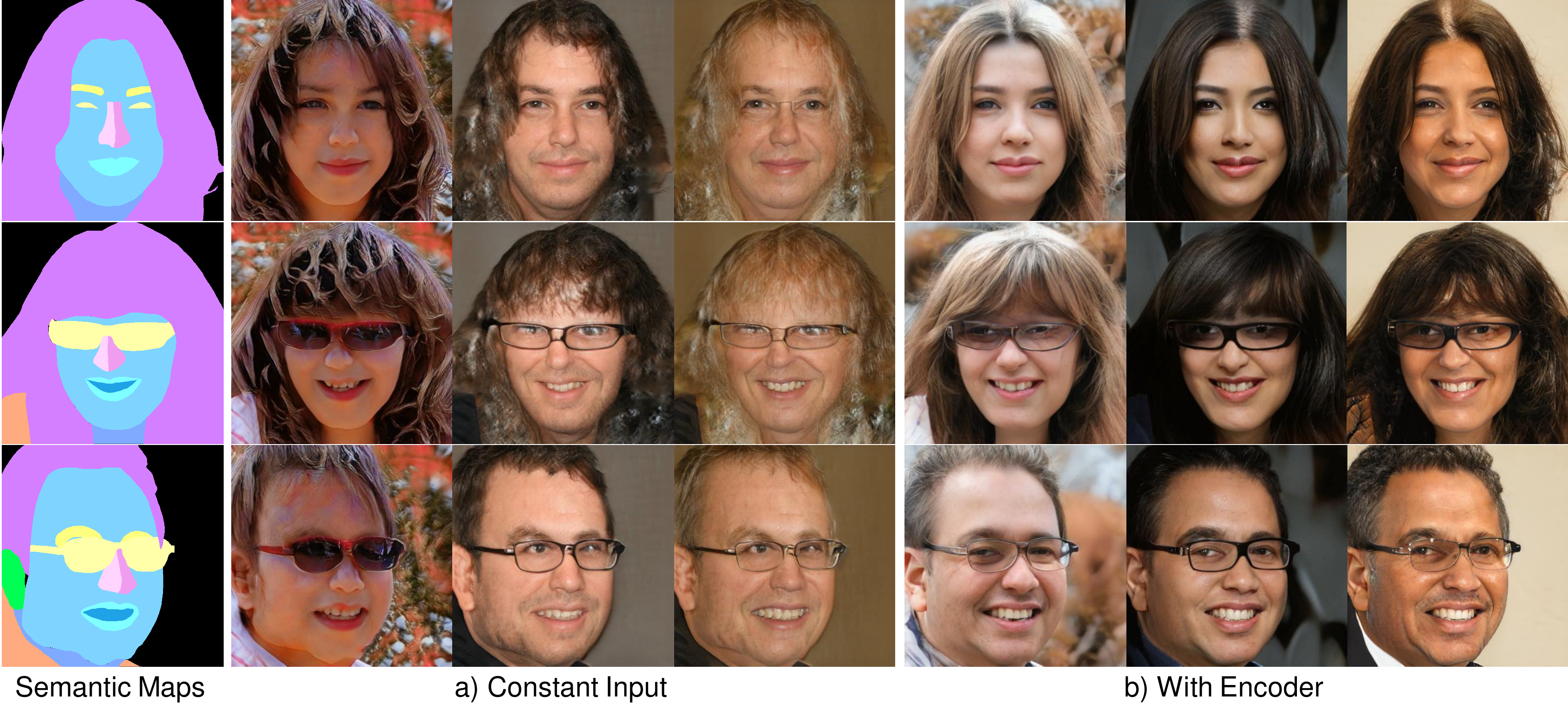}
\end{center}
  \caption{\revised{Ablation analysis on w/ and w/o the segmentation map encoder.} Each column shares the same style code while each rows denotes different segmap encoder: a) $ 512 \times 4 \times 4 $ constant input, b) $17 \times 128 \times 128 $ one hot semantic maps downscaled to $128 \times 16 \times 16$ with SIW StyleConv block, which can dynamically adjust gender styles according to different input segmentation maps and reduce gender entanglement.}
\label{fig:encoder}
\end{figure*}

\begin{figure*}[t]
\begin{center}
  \includegraphics[width=0.83\linewidth]{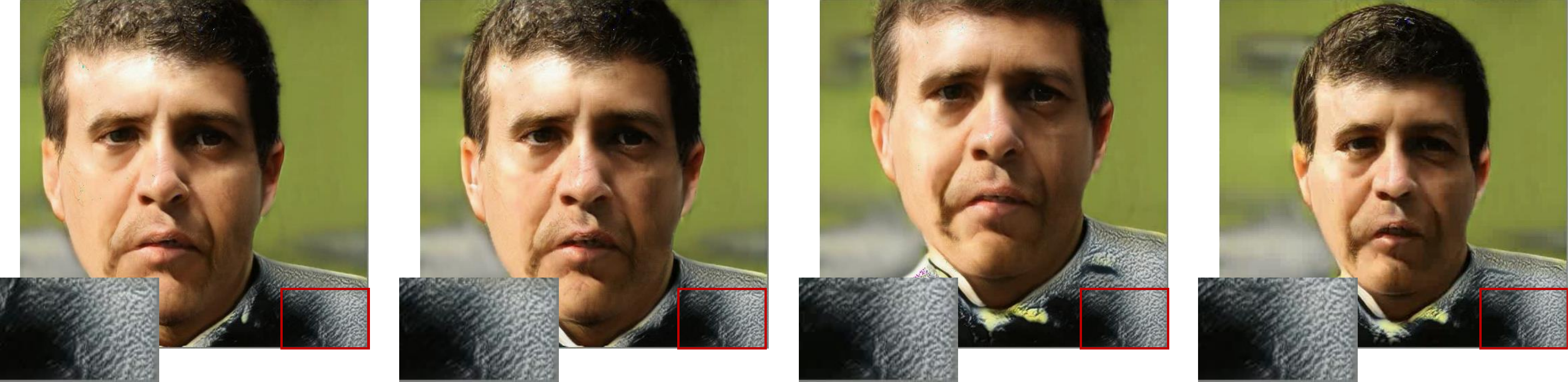}
\end{center}
  \caption{Visualization of the "phase" artifact: strong localized texture appearance keeps static during animation.}
\label{fig:phase}
\end{figure*}

\begin{figure*}[t]
\begin{center}
  \includegraphics[width=0.83\linewidth]{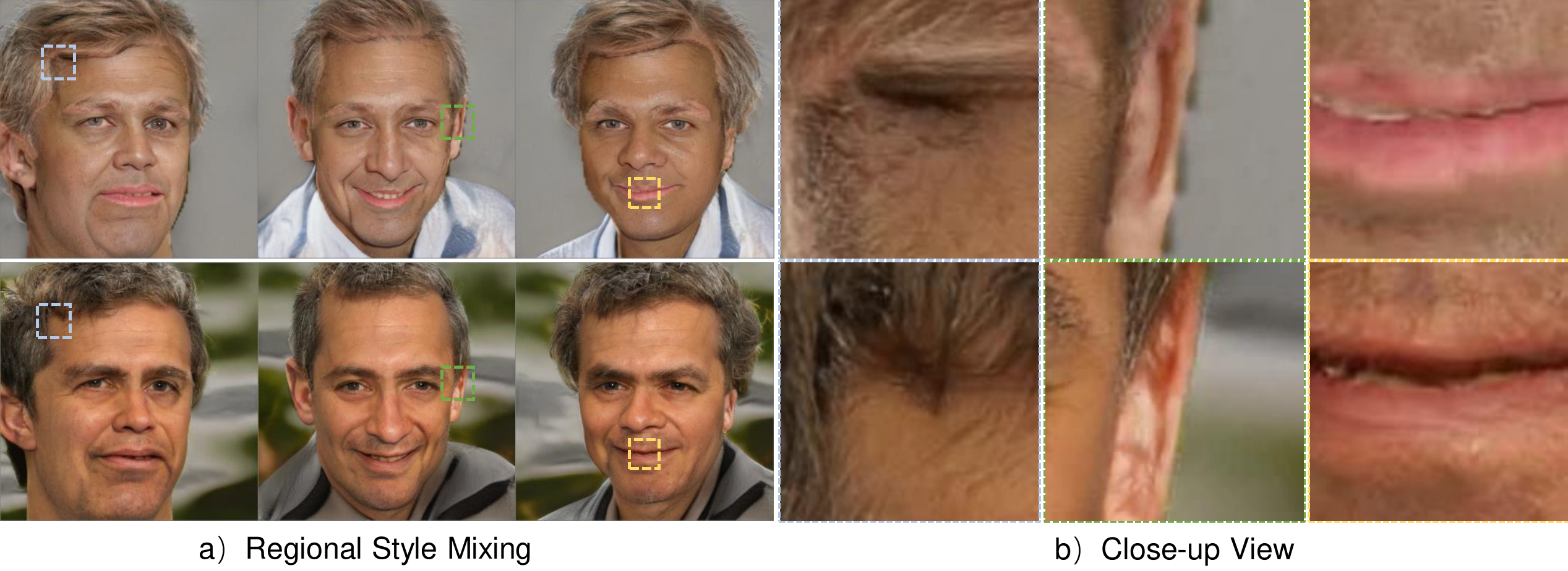}
\end{center}
  \caption{\revised{Ablation analysis on  w/ and w/o  the mixed style training scheme.} These results are generated after $800kimg$ iterations. Top row: mixing style with SIW mix style blocks. Bottom row: mixing style results without SIW mix style blocks. Hair, eyebrows and lips share the same style, while skin uses another one. Mixed style training scheme can significantly improve the smoothness around the semantic boundaries.}
\label{fig:mix_style}
\end{figure*}

\textbf{Constant Input vs. With Encoder}
\label{sec:encoder}

Unlike the StlyeGAN2, which focuses on synthesizing realistic static images, our generator attempts to enable controllable generation including local and global styling, free-viewpoint generation, and stylizing image sequences. We observe that using constant input would cause two artifacts. 1) The "phase" artifacts: as shown in Fig.~\ref{fig:phase}, refer to a strong localized appearance which is especially noticeable when viewed as an image sequence (keep static during animation). 2) The style "entanglement" artifacts: our generation process is controlled \minorrev{by} both the style codes and the semantic maps, we observe that they are usually incomparable to each other, e.g., feeding a woman's semantic map and a man's texture styles would lead to significant entanglement (shown in Fig.~\ref{fig:encoder} (a)). 

\revised{
More specifically, we observe that the "phase" artifacts are caused by the translation invariance property of CNN layer, especially deconvolution from the constant feature blocks. This results in same feature values.
At the same time, since the normalization layers are very shallow (two convolutional layers) and have a small receptive field, the design with constant input could not capture the global shape features of the semantic maps and leads to the style "entanglement" artifact.
To address these two artifacts, we use additional resBlocks to enhance the connection between style and semantic maps, as shown in Fig.~\ref{fig:encoder} (b). The encoder can efficiently reduce the artifacts by dynamically changing the generator's conditional features. In this way, the network is able to capture global shape features (e.g., contour, position in the image, etc.) and strengthen semantic control.}

\textbf{With vs. w/o SIW Style Mixing} 

To enable the regional stylizing effect, we proposed a mixed style training strategy in Sec.~\ref{sec:siw stylegan}. To demonstrate the effectiveness of this strategy, we train another model that removes this design during training and generates an image with two random styles only during evaluation (shown in the second row of Fig.~\ref{fig:mix_style}). In the case of Fig.~\ref{fig:mix_style}, the skin region is generated with style $z^{t_0}$, while other regions are synthesised with $z^{t_1}$. We can observe that significant artifacts exist on the boundaries and the images look unnatural for the second row, while the results trained with the mixed style strategy (the first row) produce soft boundaries.

\begin{figure*}[ht]
\begin{center}
  \includegraphics[width=1.0\linewidth]{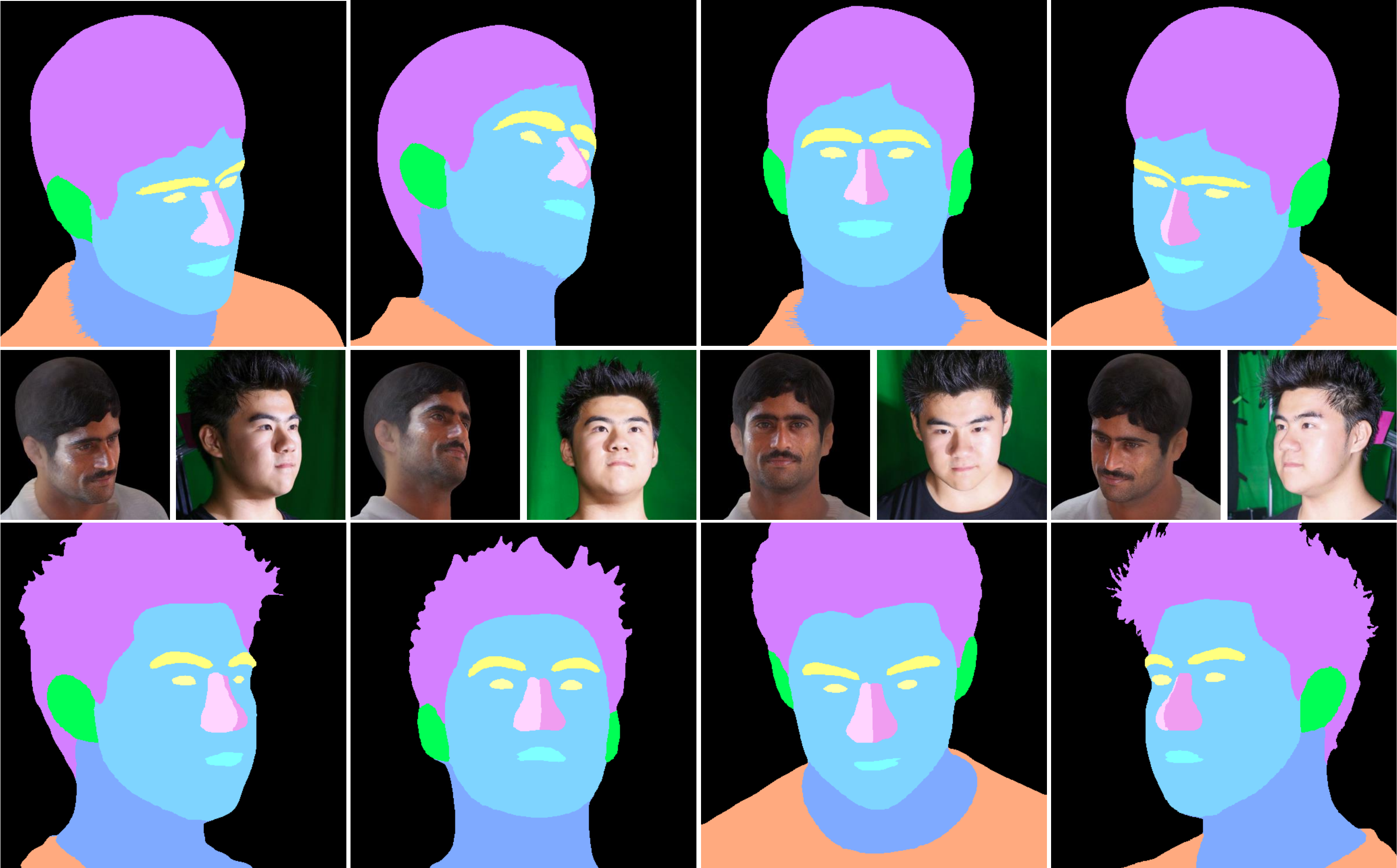}
\end{center}
  \caption{\revised{Data samples from \textit{SOF} training set. Top row: multi-view segmaps rendered from synthetic portraits (middle left). Bottom row: segmaps parsed from multi-view photos (middle right).}}
\label{fig:sof_train_data}
\end{figure*}

\section{Support Figures}
\begin{figure}[ht]
\begin{center}
  \includegraphics[width=1.0\linewidth]{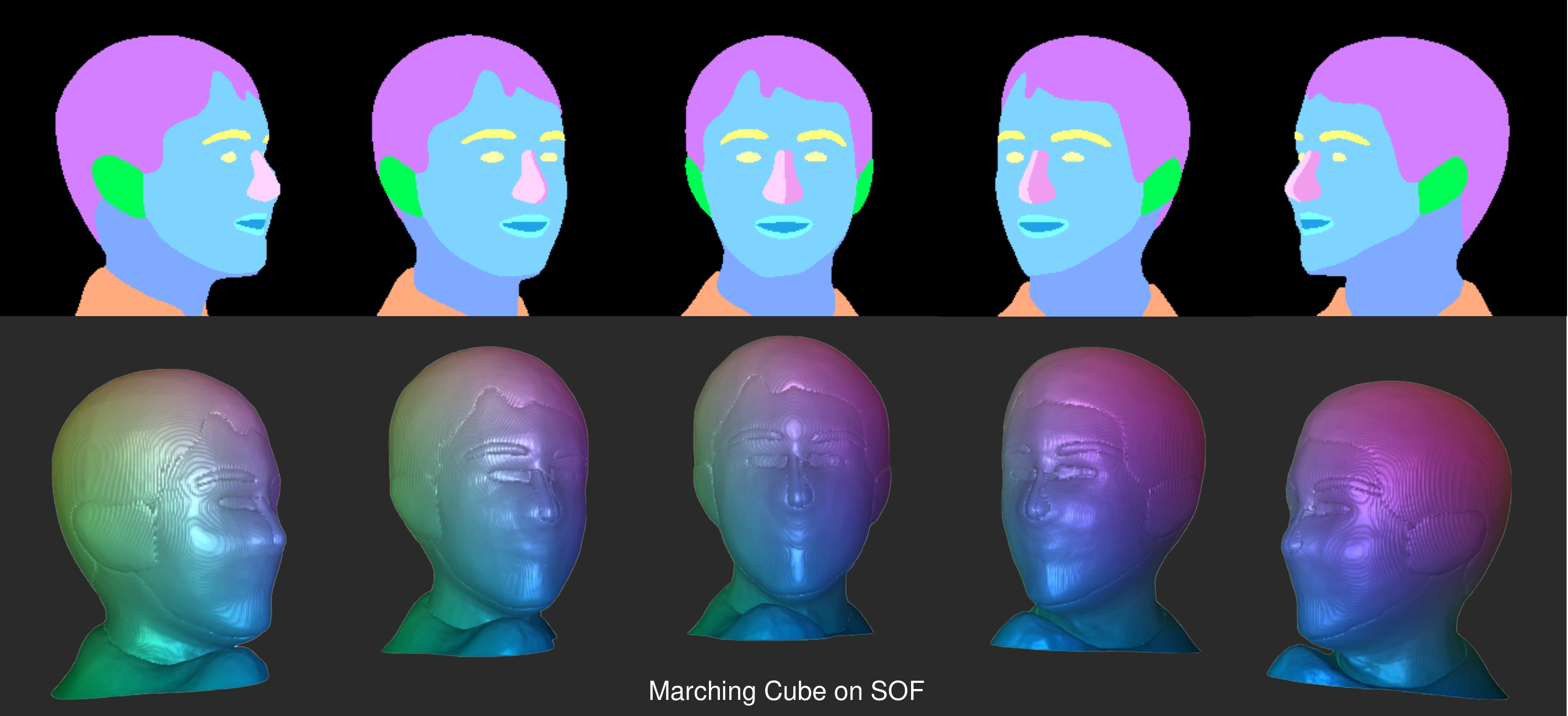}
\end{center}
  \caption{Visualization of the \textit{SOF}. To visualize the trained semantic occupancy field, we uniformly sample the occupancy of the "background" class in the \textit{SOF} and use the Marching Cubes algorithm with $level=0.5$ to obtain 3D surface (foreground). We can observe explicit boundaries around each semantic class on the surface as the \textit{SOF} encodes geometry along with its semantic properties.}
\label{fig:vis_sof}
\end{figure}

\begin{figure*}[t]
\begin{center}
  \includegraphics[width=0.98\linewidth]{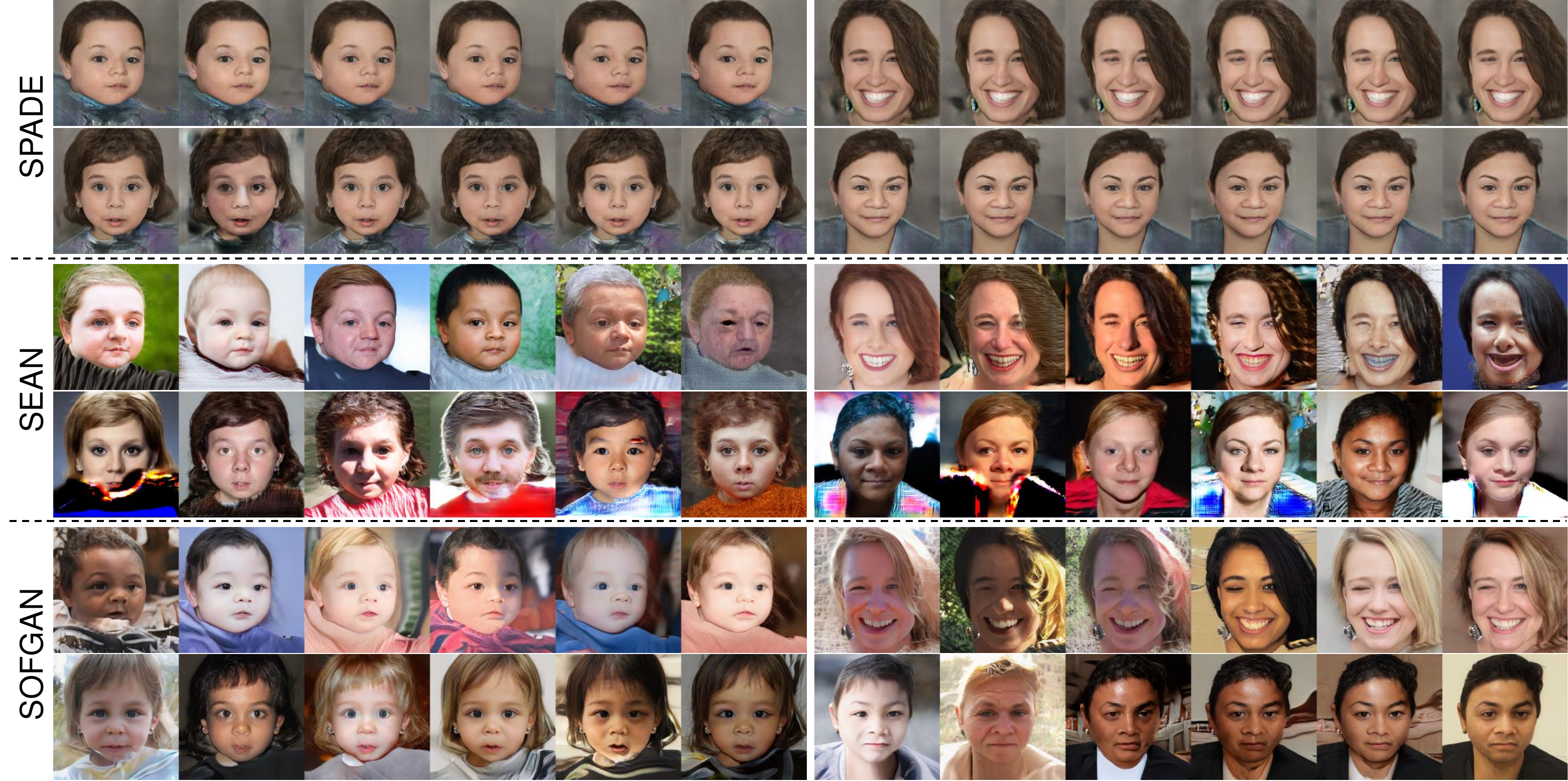}
\end{center}
\caption{\revised{Visual comparison of generation conditioned on same segmaps. Here we use the first four segmaps in the FFHQ dataset. Please refer to Sec. \ref{sec: quantitative} for the experiment setting. The corresponding quantitative results are shown in Tab. \ref{tb:fvv_eval}. Note that the models in this comparison are trained for \textbf{only} $800kimg$ iterations and we set truncation to $1$, thus we can observe some artifacts in the results.}}
\label{fig:ssg_comp}
\end{figure*}

Here we provide additional supporting figures mentioned in the main paper. Fig.~\ref{fig:sof_train_data} shows two \minorrev{multi-view} segmap examples of our \textit{SOF} training set. Fig.~\ref{fig:vis_sof} gives visualization of a trained \textit{SOF}. Fig.~\ref{fig:ssg_comp} compares the controlled image generation conditioned on same segmaps.

\section{Additional Results}
\begin{figure*}[t]
\begin{center}
   \includegraphics[width=\linewidth]{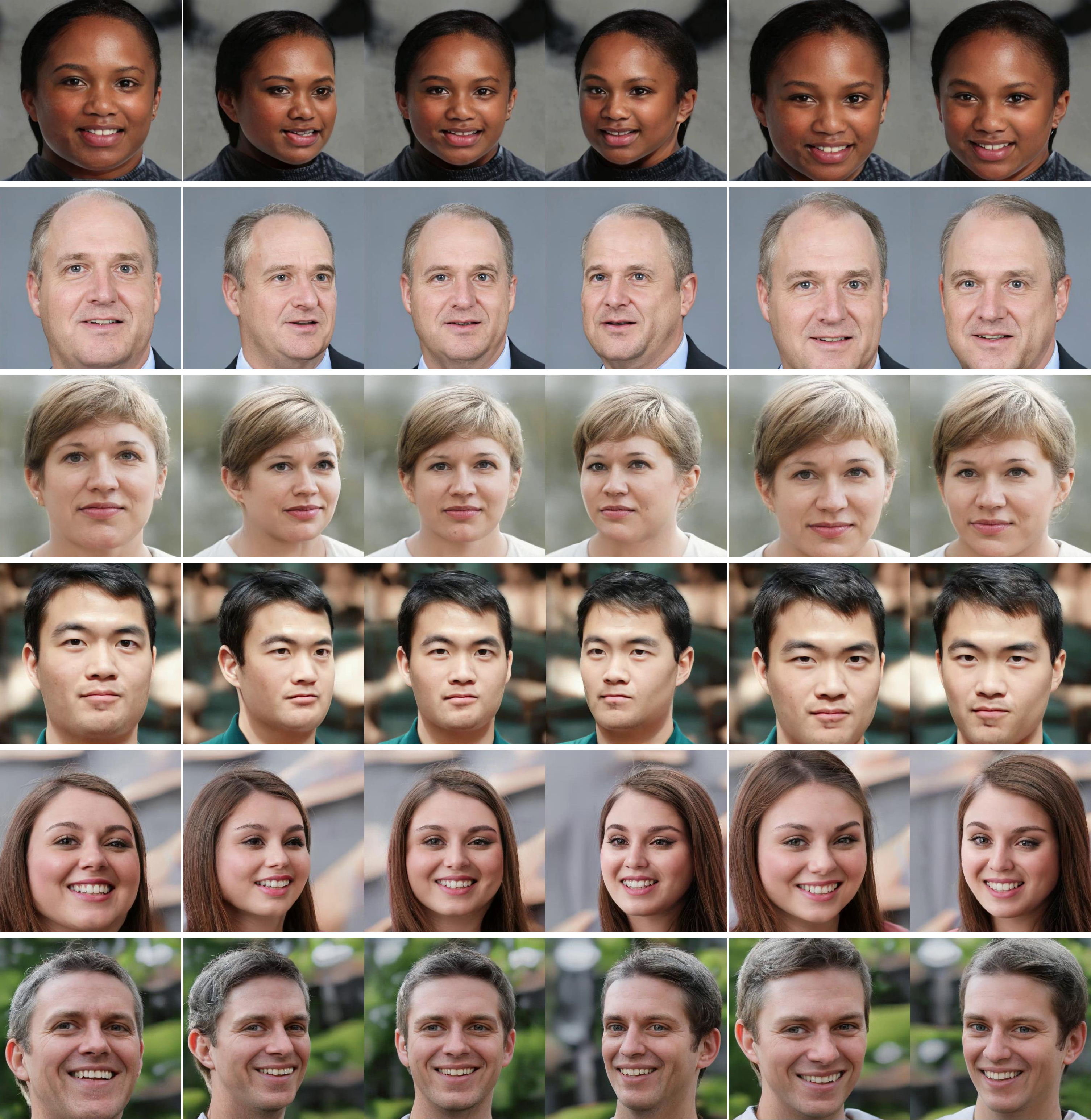}
\end{center}
   \caption{\revised{Free-viewpoint generation results. Images in the same row share the same geometry and texture style vectors, but use different camera poses. We can observe that our method is able to preserve shape and texture consistency even under large view angle variation.}}
\label{fig:free-viewpoint}
\end{figure*}

\begin{figure*}[t]
\begin{center}
  \includegraphics[width=0.95\linewidth]{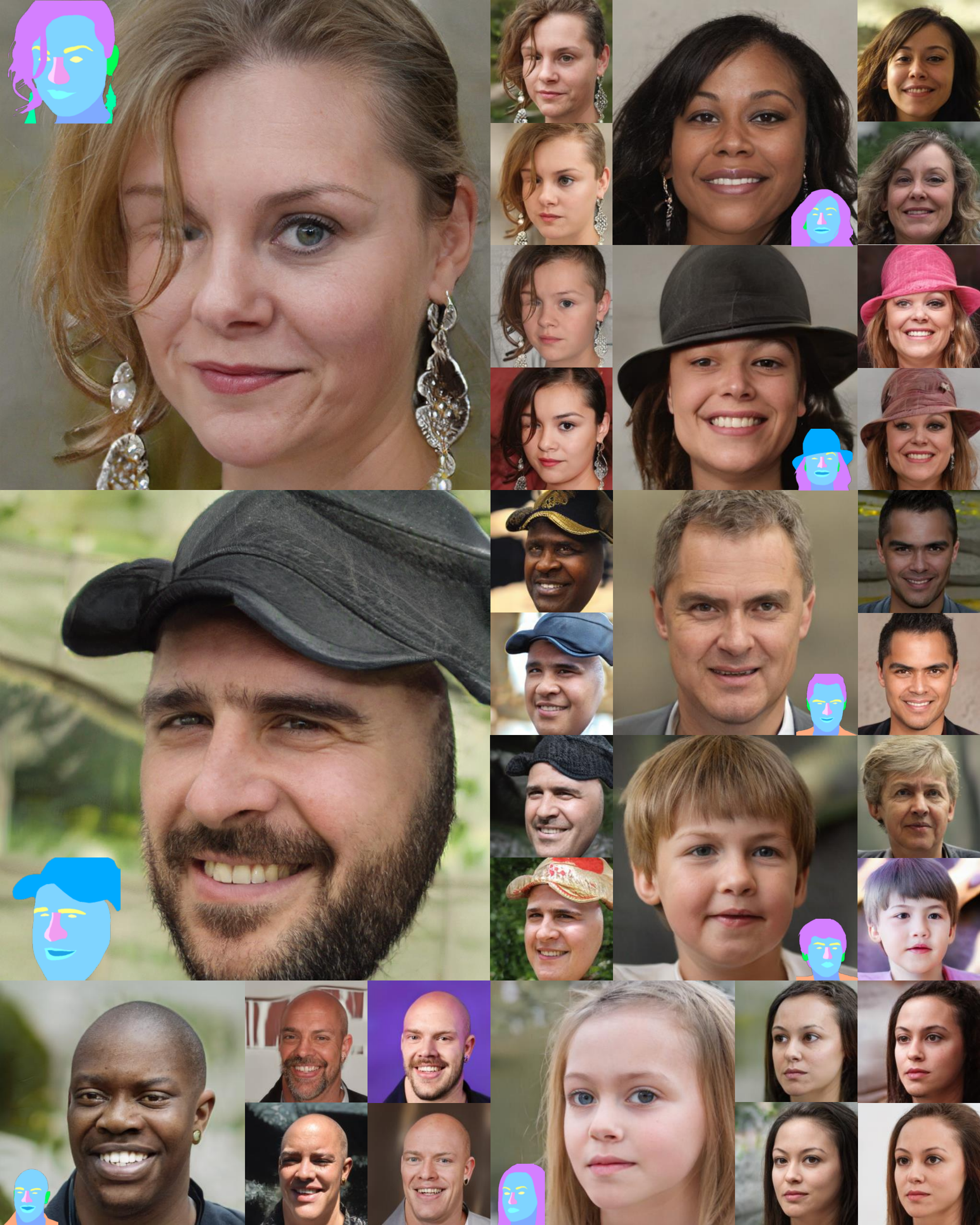}
\end{center}
\caption{Results for global style adjustment.}
\label{fig:global_style_1}
\end{figure*}

\begin{figure*}[t]
\centering
    \includegraphics[width=1.0\linewidth]{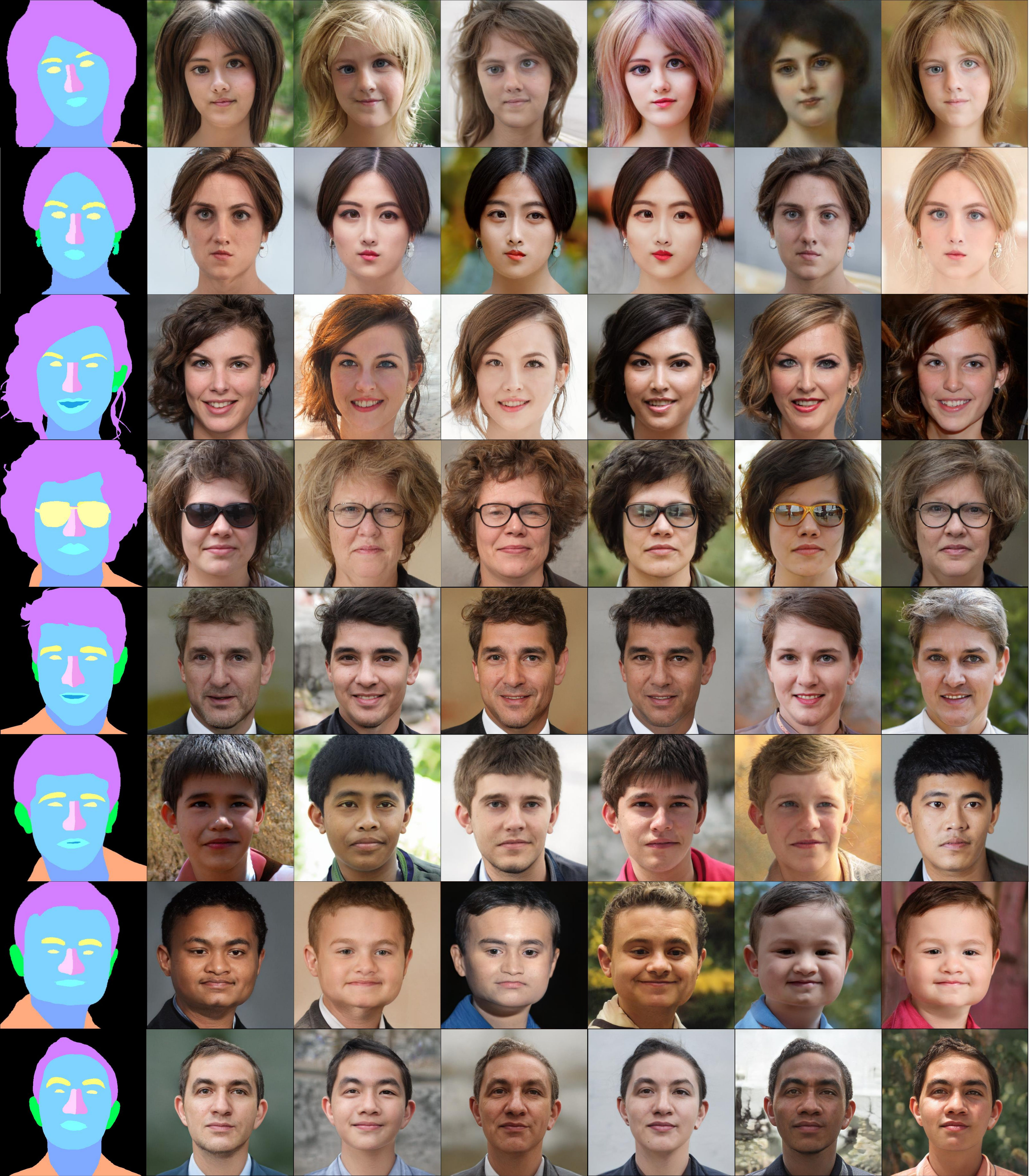}
\caption{Results for global style adjustment.}
\label{fig:global_style_2}
\end{figure*}

\begin{figure*}[t]
\begin{center}
  \includegraphics[width=0.98\textwidth]{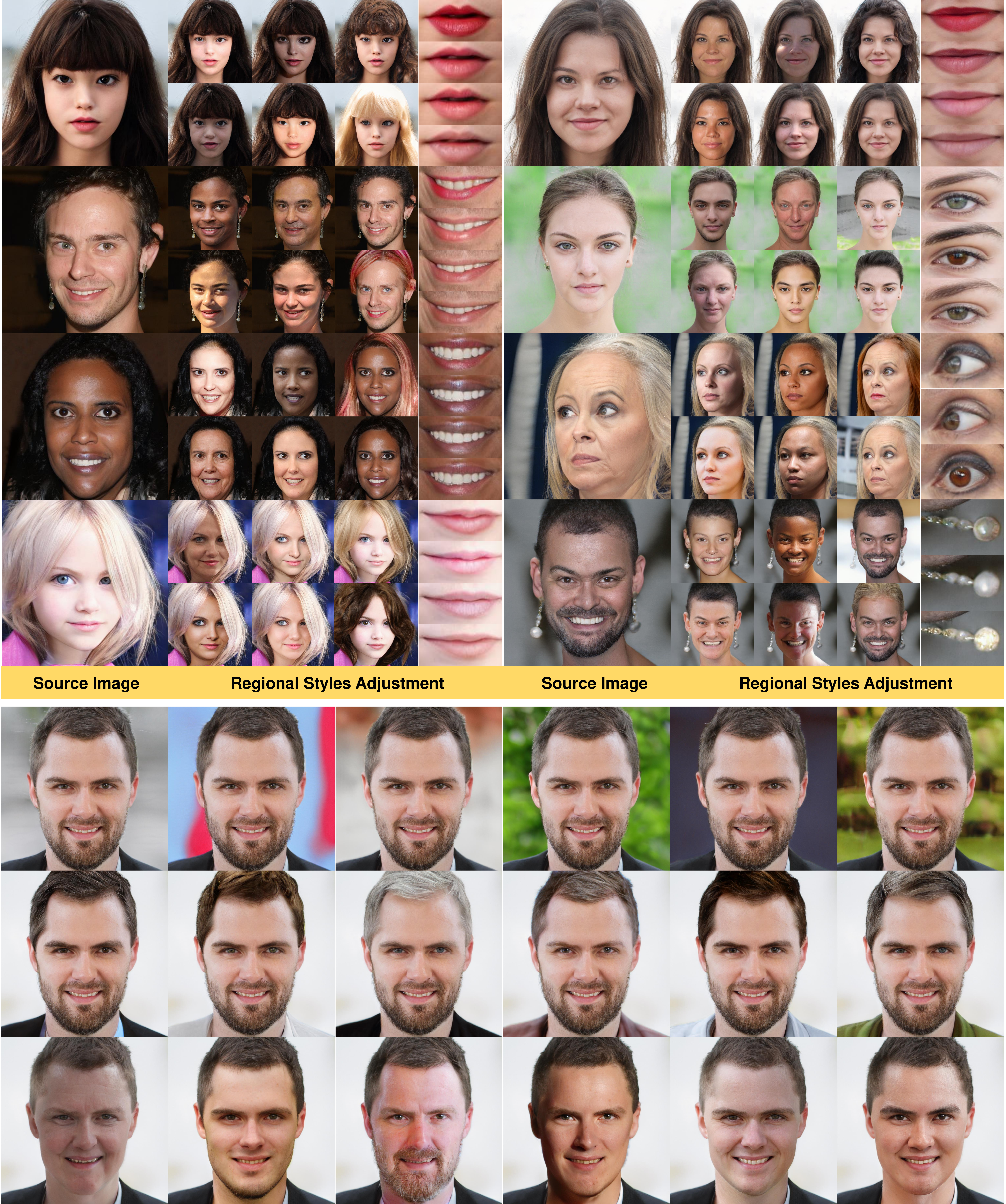}
\end{center}
\caption{Results for regional style adjustment.}
\label{fig:regional_styles}
\end{figure*}

\begin{figure*}[t]
\begin{center}
  \includegraphics[width=1.0\linewidth]{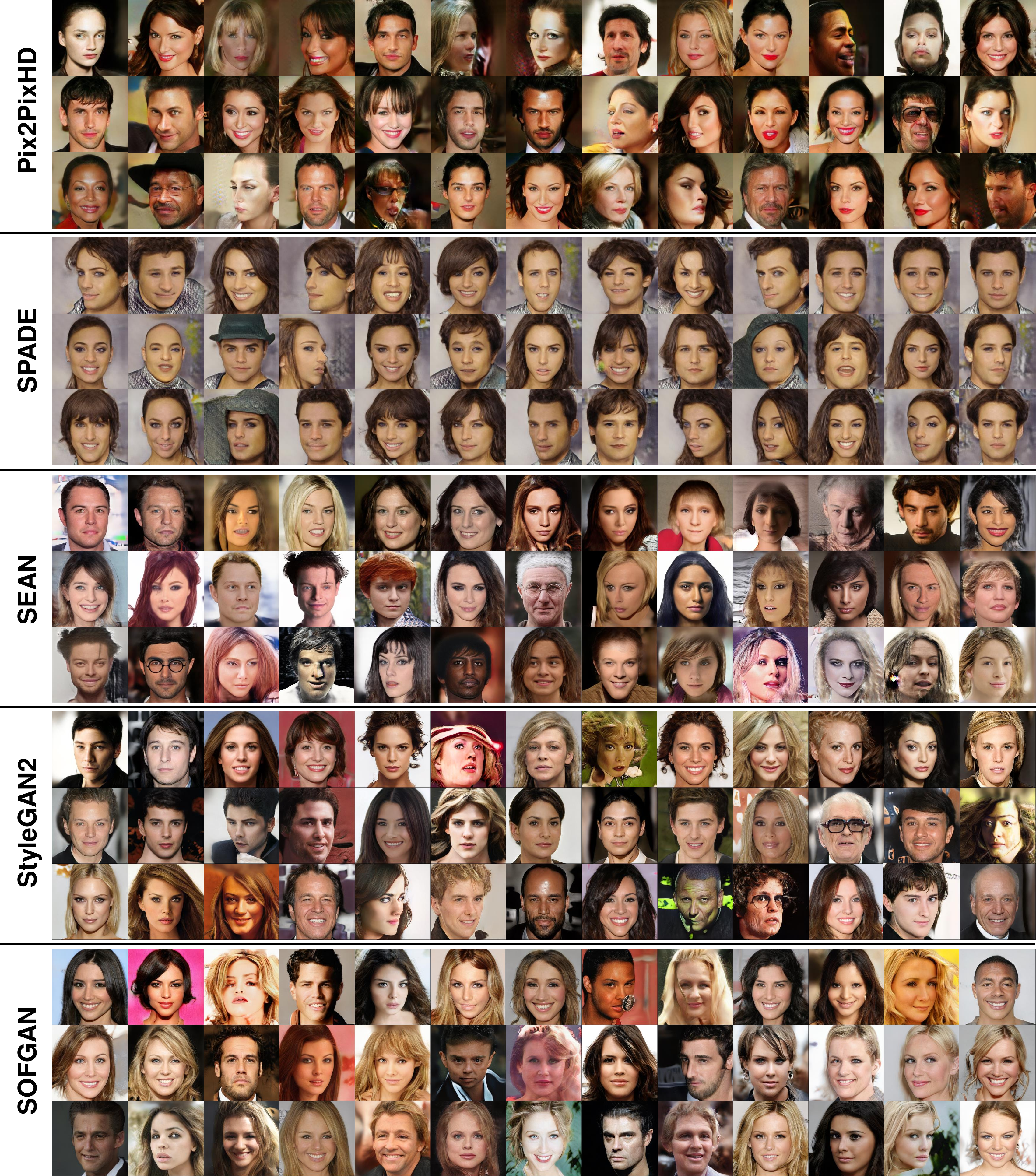}
\end{center}
\caption{Visual comparison on CelebAMask-HQ dataset ~\cite{CelebAMask-HQ}. Each model is trained with $800k$ images in resolution $512^2$.}
\label{fig:comparision_celeba}
\end{figure*}

\begin{figure*}[t]
\begin{center}
  \includegraphics[width=1.0\linewidth]{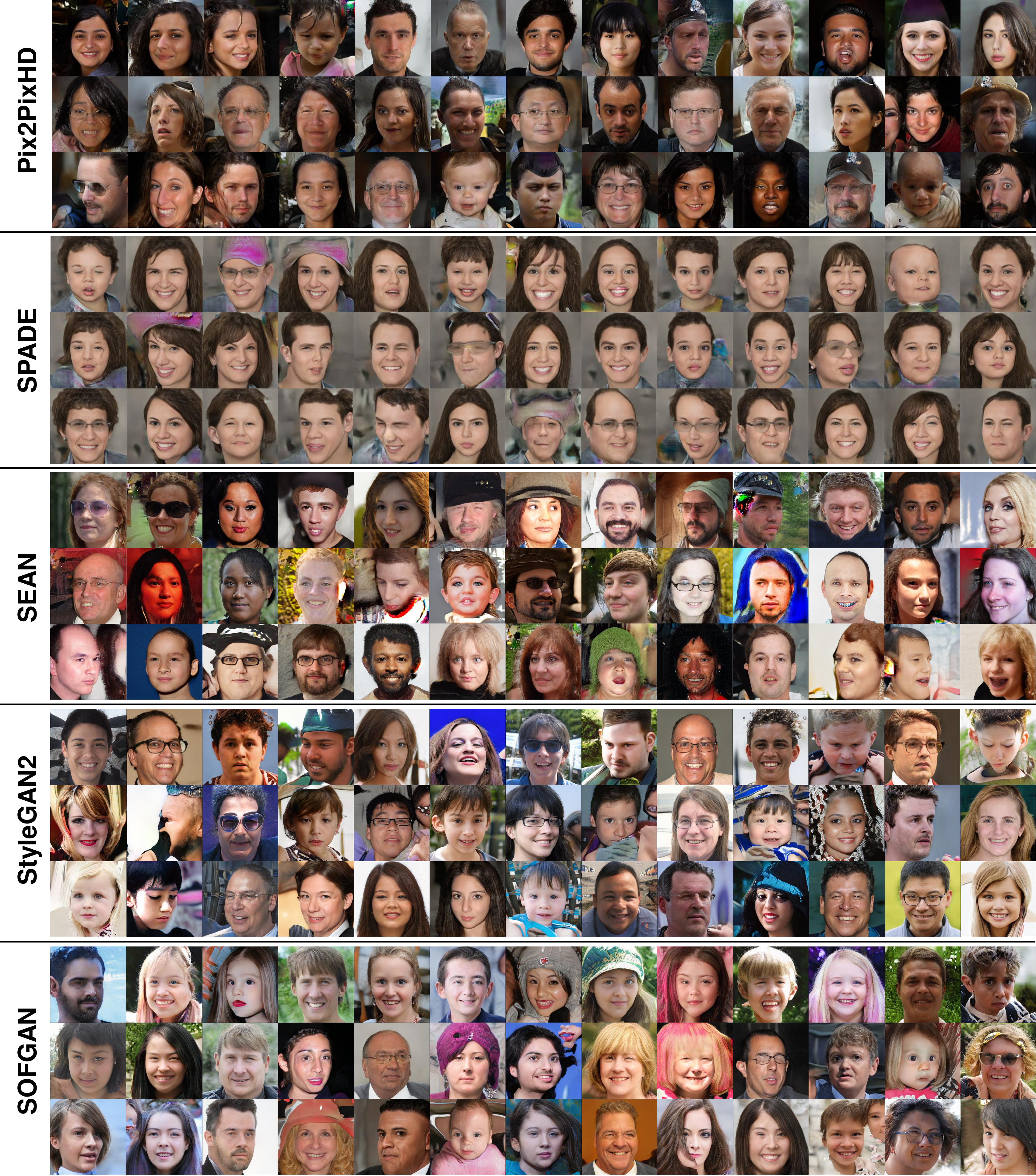}
\end{center}
\caption{Visual comparison on FFHQ dataset ~\cite{karras2019style}. Each model is trained with $800k$ images in resolution $512^2$.}
\label{fig:comparision_ffhq}
\end{figure*}

We provide additional results to better demonstrate the robustness of our SofGAN. Fig.~\ref{fig:free-viewpoint} shows free-viewpoint generation results conditioned on different \textit{SOF}.  Fig.~\ref{fig:global_style_1}, \ref{fig:global_style_2} and \ref{fig:regional_styles} show style adjustment on both global and local semantic regions. 
Fig.~\ref{fig:comparision_celeba} and \ref{fig:comparision_ffhq} demonstrate the visual comparisons between Pix2PixHD~\cite{wang2018pix2pixHD}, SPADE~\cite{park2019SPADE}, SEAN~\cite{zhu2020sean}, StyleGAN2~\cite{karras2019analyzing} and our SofGAN. For the corresponding quantitative evaluation and implementation details, please refer to Sec.~\ref{sec: quantitative} and \ref{sec: implementation}.

\begin{figure*}[t]
\begin{center}
  \includegraphics[width=0.98\linewidth]{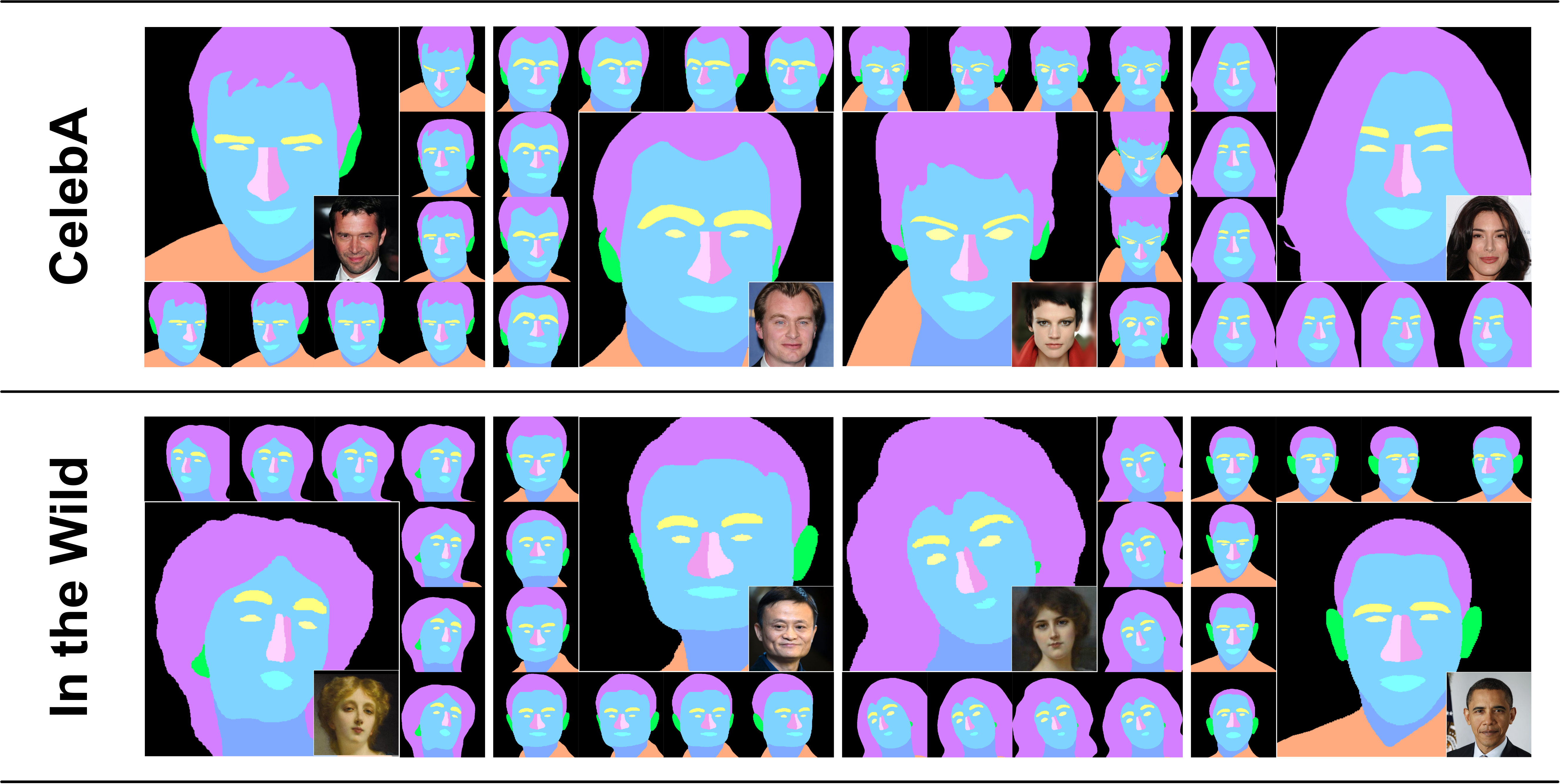}
\end{center}
  \caption{\revised{Generating multi-view segmaps for real images. We first parse a monocular segmap from a given image, then project the segmap (central segmap) to SOF and generate multi-view segmaps (surrounding segmaps).}}
\label{fig:seg_mv}
\end{figure*}

\begin{figure*}[t]
\begin{center}
  \includegraphics[width=0.97\textwidth]{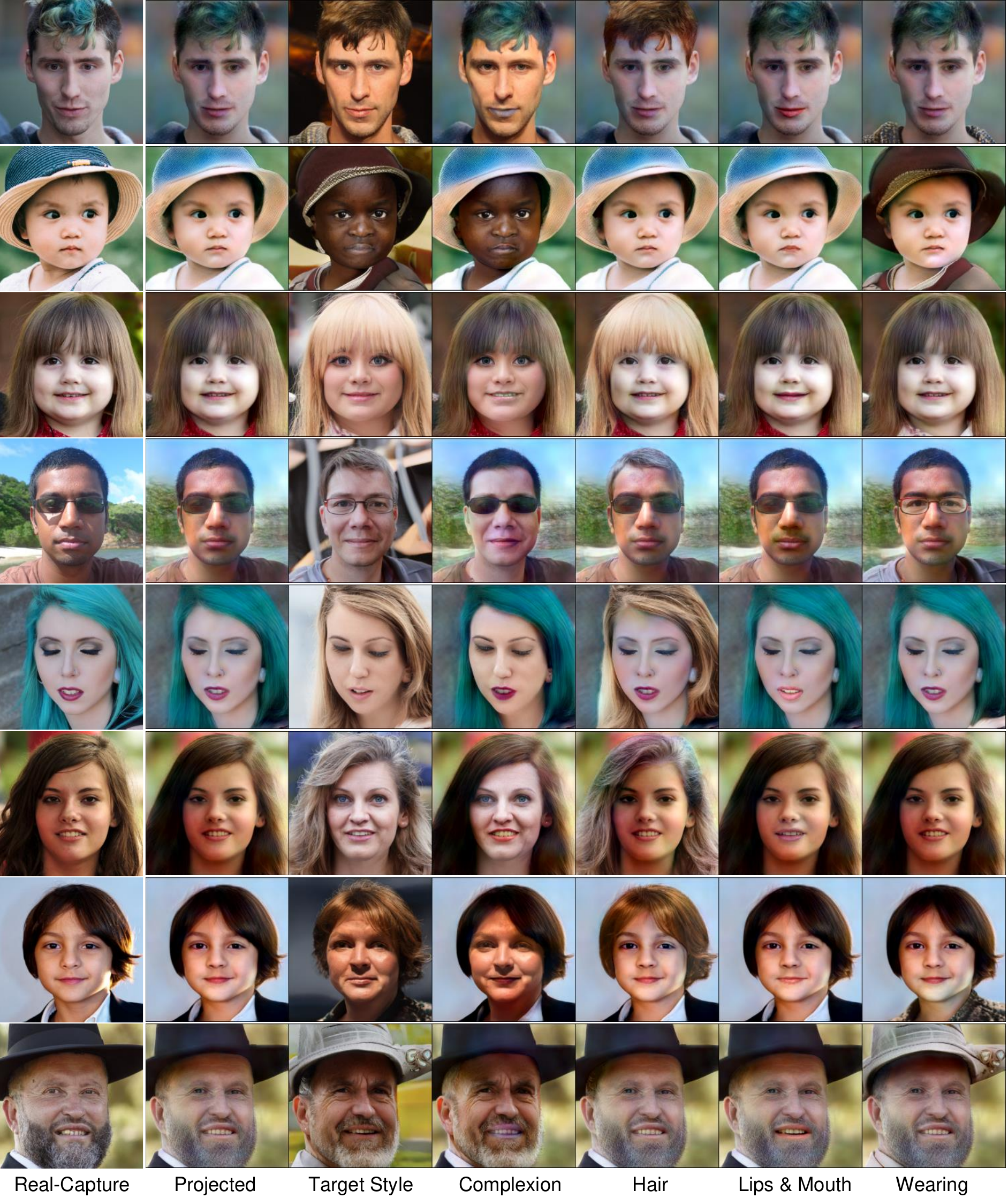}
\end{center}
\caption{\revised{Results for real-photo editing. We first project the real-captured photos into our texture space $z^{t_0}$, then we randomly sample a target texture style $z^{t_1}$ from our texture space, furthermore, we regionally edit the project image via the regional style mixing scheme mentioned in Sec. ~\ref{sec:applications}.}}
\label{fig:real_editing}
\end{figure*}

\end{document}